\documentclass[11pt]{article}

\usepackage[preprint]{acl}

\usepackage{times}
\usepackage{latexsym}

\usepackage[T1]{fontenc}

\usepackage[utf8]{inputenc}

\usepackage{microtype}

\usepackage{inconsolata}

\usepackage{graphicx}
\usepackage{amsmath}
\usepackage{amssymb}
\usepackage[table]{xcolor}
\usepackage{subcaption}
\usepackage{booktabs}
\usepackage{multirow}
\usepackage{algorithm}
\usepackage{algorithmic}
\usepackage[most]{tcolorbox}
\title{What If Consensus Lies? Selective-Complementary\\ Reinforcement Learning at Test Time}

\author{%
  Dong Yan\textsuperscript{1,2}, Jian Liang\textsuperscript{1,2}\thanks{Corresponding Author}, Yanbo Wang\textsuperscript{1,2}, Shuo Lu\textsuperscript{2}, Ran He\textsuperscript{1,2}, Tieniu Tan\textsuperscript{1,2,3}\\
    \textsuperscript{1}School of Artificial Intelligence, University of Chinese Academy of Sciences\\
    \textsuperscript{2}NLPR \& MAIS, Institute of Automation, Chinese Academy of Sciences\\
    \textsuperscript{3}Nanjing University\\
    \texttt{yandong2025@ia.ac.cn, liangjian92@gmail.com}
}

\begin{document}
 \maketitle
\begin{abstract}
Test-Time Reinforcement Learning (TTRL) enables Large Language Models (LLMs) to enhance reasoning capabilities on unlabeled test streams by deriving pseudo-rewards from majority voting consensus.
However, existing TTRL methods rely exclusively on positive pseudo-labeling strategies.
Such reliance becomes vulnerable under challenging scenarios where answer distributions are highly dispersed, resulting in weak consensus that inadvertently reinforces incorrect trajectories as supervision signals.
In this paper, we propose SCRL (Selective-Complementary Reinforcement Learning), a robust test-time reinforcement learning framework that effectively mitigates label noise amplification.
SCRL develops Selective Positive Pseudo-Labeling, which enforces strict consensus criteria to filter unreliable majorities. 
Complementarily, SCRL introduces Entropy-Gated Negative Pseudo-Labeling, the first negative supervision mechanism in TTRL, to reliably prune incorrect trajectories based on generation uncertainty. 
Extensive experiments on multiple reasoning benchmarks demonstrate that SCRL achieves substantial improvements over baselines, while maintaining robust generalization and training stability under constrained rollout budgets.
Our code is available at \href{https://github.com/Jasper-Yan/SCRL}{https://github.com/Jasper-Yan/SCRL}.
\end{abstract}

\section{Introduction}
Reinforcement Learning with Verifiable Rewards (RLVR) \citep{jaech2024openai,shao2024deepseekmath,yang2025qwen3} has significantly advanced the reasoning capabilities of Large Language Models (LLMs) , enabling state-of-the-art performance in verifiable domains such as mathematics and coding \citep{gao2024designing,setlur2024rl,wang2024enhancing}. 
Guided by ground-truth labels or rule-based verification signals, RLVR allows policy optimization to directly reinforce trajectories that lead to correct outcomes. 
However, the reliance on extensive manually-annotated data creates a fundamental limitation: as task complexity and diversity grow, acquiring high-quality supervision becomes increasingly difficult. 
To bridge this gap, Test-Time Reinforcement Learning (TTRL) has emerged as a critical paradigm for unsupervised reasoning \citep{zuo2025ttrl,yang2025spell,jayalath2025compute,yuan2025wisdom}. 
TTRL allows models to self-improve on unlabeled test streams by generating diverse rollouts and leveraging majority voting consensus to derive pseudo-rewards for policy updates.

\begin{figure}[t]
  \includegraphics[width=\columnwidth]{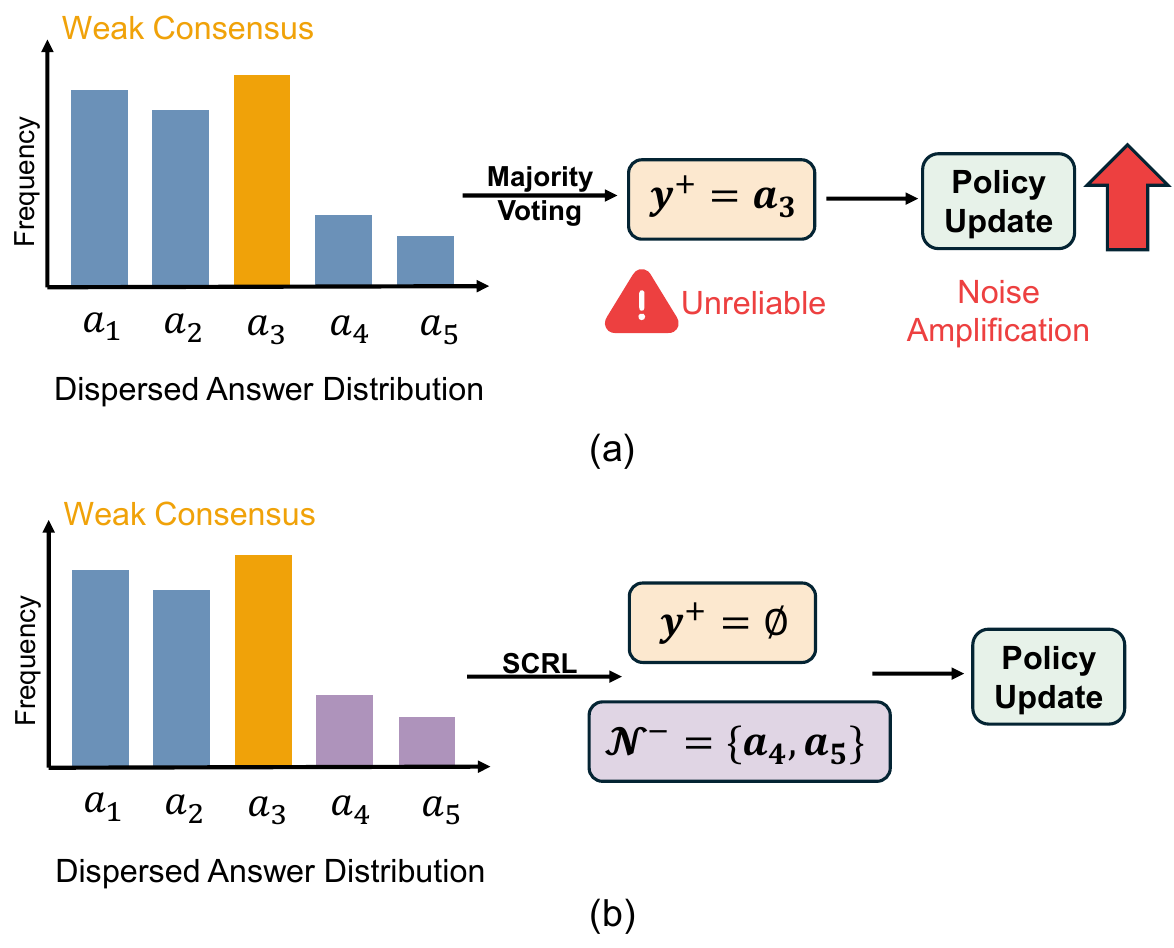}
  \caption{Comparison of pseudo-labeling strategies under weak consensus. (a) Majority voting assigns the positive label despite dispersed answer distribution. (b) SCRL abstains from positive labeling when consensus is insufficient and identifies negative labels.}
  \label{fig:intro}
\end{figure}

While TTRL offers a promising direction for unsupervised reasoning, existing methods \citep{zuo2025ttrl,yu2025restrain,wen2025self,wang2025self} that rely on majority voting and its variants, such as soft-weighted consensus and self-play mechanisms, face inherent limitations rooted in their exclusive focus on positive pseudo-labeling. 
These methods require substantial rollout budgets to achieve reliable consensus; however, on challenging problems, the answer distribution remains highly dispersed even with extensive sampling. 
As shown in Figure~\ref{fig:intro}(a), this dispersion weakens the consensus, which may result in incorrect trajectories being utilized as supervision signals \citep{stahlberg2022uncertainty,liu2025ettrl}. 
Consequently, the model prematurely converges toward spurious solutions through policy optimization \citep{shi2025heimdall,huang2024mirror}.
Note that when rollout budgets are constrained, this vulnerability is particularly pronounced, where insufficient sampling coverage increases consensus instability.
In addition, while identifying a correct answer is difficult under high uncertainty, recognizing incorrect answers is comparatively reliable.
Nevertheless, existing methods overlook the potential of negative labeling. 
As illustrated in Figure~\ref{fig:intro}(b), when credible positive consensus is absent, a robust strategy is to employ negative labels to prune the search space, which allows the model to eliminate errors and update toward more promising regions without prematurely committing to any single answer.

To address these critical issues, we propose \textbf{SCRL} (\textbf{S}elective-\textbf{C}omplementary \textbf{R}einforcement \textbf{L}earning), a robust framework that effectively mitigates label noise amplification in unsupervised test-time reinforcement learning. 
SCRL develops \textit{Selective Positive Pseudo-Labeling} which enforces strict consensus and margin criteria, ensuring that positive supervision is only provided when the answer distribution exhibits sharp concentration and clear separation from alternatives. 
This mechanism can prevent the amplification of unreliable majorities when the answer distribution is dispersed. 
Complementing this, SCRL introduces \textit{Entropy-Gated Negative Pseudo-Labeling}, the first mechanism in test-time reinforcement learning that integrates negative supervision to identify and penalize incorrect trajectories. 
By isolating answers that exhibit both low frequency and high uncertainty, the model reliably prunes implausible solutions without eliminating potentially correct low-frequency answers. 
To calibrate the reinforcement magnitude based on consensus strength, we design \textit{Dynamic Reward Shaping} that integrates credible positive signals with informative negative signals, enabling SCRL to maintain exploration capacity while systematically narrowing the search space and achieve robust unsupervised reinforcement learning.

Extensive experiments on multiple reasoning benchmarks consistently demonstrate that SCRL significantly outperforms baseline methods, particularly on challenging problems and under constrained rollout budgets. 
Our contributions can be summarized as follows:
\begin{itemize} 
\item We propose SCRL, a test-time reinforcement learning framework that mitigates label-noise amplification under weak consensus.
\item SCRL incorporates strict consensus criteria to filter unreliable majorities, restricting positive supervision to concentrated distributions.
\item SCRL introduces negative supervision in test-time reinforcement learning for the first time, which eliminates implausible trajectories without discarding potentially valid rare solutions.
\item Extensive experiments consistently demonstrate that SCRL outperforms baselines particularly under constrained rollout budgets, while ablation studies and label-quality analyses validate the necessity of each component.
\end{itemize}

\section{Related Work}
\begin{figure*}[t]
\vspace{-1.5\baselineskip}
    \centering
    \includegraphics[width=1.0\textwidth]{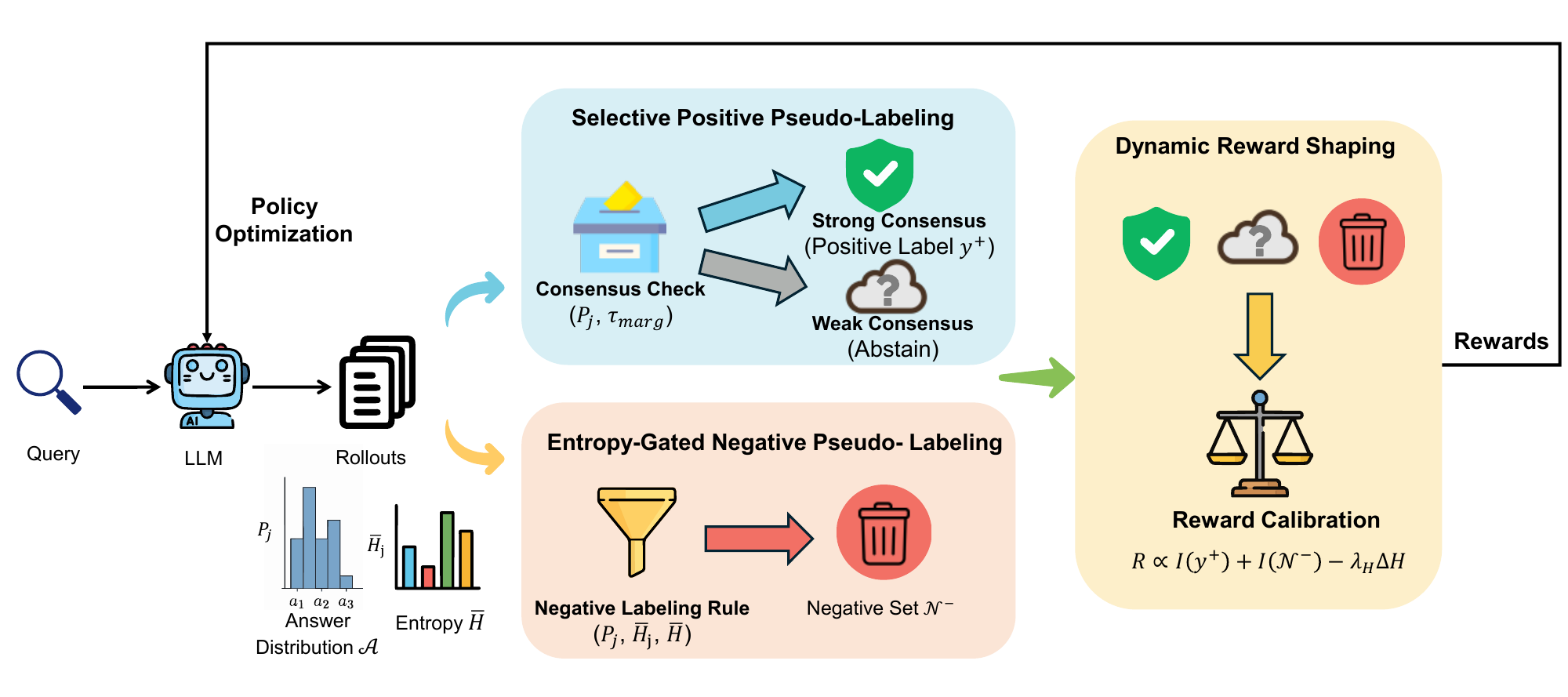}  
    \caption{Overview of the SCRL framework. SCRL addresses test-time label noise through three components: selective positive pseudo-labeling enforces strict consensus thresholds to prevent reinforcing unreliable majorities; entropy-gated negative pseudo-labeling identifies negative labels by isolating answers that are both rare and exhibit high uncertainty, pruning the search space without eliminating valid candidates; dynamic reward shaping constructs distribution-aware rewards that scale with consensus strength and penalize uncertainty trajectories.}
    \label{fig:overview}
\end{figure*}

\subsection{RL for Reasoning}
Reinforcement learning (RL) has emerged as a critical approach for enhancing the instruction-following and complex reasoning capabilities of LLMs. Reinforcement Learning from Human Feedback (RLHF) \citep{ouyang2022training} aligns base models with human preferences via annotated preference data and policy optimization methods \citep{schulman2017proximal,rafailov2023direct,meng2024simpo,cui2025process,wang2025comprehensivesurveytrustworthinessreasoning,lu2025uni}. To reduce reliance on human labels, Reinforcement Learning with Verifiable Rewards (RLVR) \citep{shao2024deepseekmath,yu2025dapo,feng2025don,lu2025deepresearch} replaces preference rewards with verifiable signals, enabling objective and automated evaluation which has proven especially effective in math and code domains \citep{yang2025qwen3,lambert2024tulu,wang2025reinforcement}.
Furthermore, Reinforcement Learning from Internal Feedback (RLIF) \citep{zhao2025learning,shafayat2025can,prasad2024self,liu2025understanding} derives intrinsic rewards from the model’s confidence, entropy or self-consistency across its reasoning paths \citep{tan2025diagnosing,zhang2025right,zhang2025consistent,zhao2025absolute,yan2025mission}. For example, Intuitor \citep{zhao2025learning} utilizes the model's confidence as a sparse intrinsic reward to reinforce high confidence reasoning paths, while EMPO \citep{zhang2025right} incentivizes reasoning by minimizing the predictive entropy of LLM outputs in a latent semantic space. 
Our work belongs to the RLIF paradigm and uniquely leverages both positive and negative signals derived from the model's output distribution to enable robust test-time reinforcement learning.

\subsection{Unsupervised Reasoning at Test Time}
Test-Time Reinforcement Learning (TTRL) has emerged as a crucial paradigm for adapting LLMs to unlabeled test streams, utilizing majority voting consensus as a verifiable pseudo-reward for online policy optimization \citep{zuo2025ttrl,wei2025unsupervised,yu2025restrain,liu2025ettrl,wang2025self,prabhudesai2025maximizing, wu2025spine,yu2026understanding,zhou2025evolving}. Recent research has focused on robust unsupervised reward estimation: RESTRAIN \citep{yu2025restrain} employs soft-weighted pseudo-labels and penalizes low-confidence responses to enhance training stability, while Self-Harmony \citep{wang2025self} utilizes a self-play mechanism to verify positive labels. SPINE \citep{wu2025spine} stabilizes training by restricting updates to high-entropy forking tokens, whereas ETTRL \citep{liu2025ettrl} enhances efficiency through entropy-fork tree rollouts to mitigate early-stage estimation bias. MM-UPT \citep{wei2025unsupervised} extends this paradigm to the multimodal domain, validating the approach for complex vision-language reasoning tasks.  However, relying solely on voting-based methods for positive label assignment can amplify noise when consensus is weak. 
Our work introduces selective positive labeling with strict consensus criteria and complements it with negative labeling to prune the search space without premature convergence.

\section{Method}
In this section, we propose \textbf{SCRL} (\textbf{S}elective-\textbf{C}omplementary \textbf{R}einforcement \textbf{L}earning), a robust framework for test-time reinforcement learning to mitigate label noise amplification in unsupervised settings. As illustrated in Figure~\ref{fig:overview}, SCRL consists of three components: Selective Positive Pseudo-Labeling (Section~\ref{sec:credible-consensus}), Entropy-Gated Negative Pseudo-Labeling (Section~\ref{sec:pseudo-negative}), and Dynamic Reward Shaping (Section~\ref{sec:reward-shaping}).

\subsection{Preliminaries}
We adopt the Grouped Relative Policy Optimization (GRPO) \citep{shao2024deepseekmath} as our main RL algorithm. For a given query $q$, the policy samples a group of $G$ responses $\mathcal{O} = \{o_1, \dots, o_G\}$ from the sampling policy $\pi_{old}$. Each response receives a reward $R_i$ and GRPO constructs a group-normalized advantage $\hat{A}_i$ shared across tokens:
\begin{equation}
\begin{split}
    \hat{A}_i &= \frac{R_i - \mu}{\sigma}.
\end{split}
\end{equation}

The parameters $\theta$ are updated by maximizing the objective:
\begin{equation}
\begin{split}
    \mathcal{J}_{GRPO}(\theta) = \mathbb{E}_{q \sim \mathcal{Q}, \mathcal{O} \sim \pi_{old}} \bigg[ \frac{1}{G} \sum_{i=1}^{G} \frac{1}{|o_i|} \sum_{t=1}^{|o_i|} \\
    \min \Big( \rho_{i,t} \hat{A}_i, \text{clip}(\rho_{i,t}, 1 - \epsilon, 1 + \epsilon) \hat{A}_i \Big) \bigg],
\end{split}
\end{equation}
where $\rho_{i,t} = \pi_{\theta}(o_{i,t} | q, o_{i,<t}) / \pi_{old}(o_{i,t} | q, o_{i,<t})$ is the importance sampling ratio.

\subsection{Selective Positive Pseudo-Labeling}
\label{sec:credible-consensus}
Majority voting assigns positive pseudo-label by selecting the most frequent answer among rollouts \citep{zuo2025ttrl,wang2025self,liu2025ettrl}. This implicitly assumes that the most frequent answer is a reliable label for the unknown ground truth. However, on difficult queries or with limited rollout budgets, the answer distribution becomes dispersed: correct trajectories are sparse and erroneous solutions are diverse. In this situation, majority voting can produce false-positive supervision by promoting a wrong answer, which is further amplified under GRPO due to group normalization. Let $f=\frac{1}{G}\sum_{i=1}^{G}R_i$ be the fraction of rollouts labeled positive within a group. The normalized advantage for positive samples is:
\begin{equation}
\hat A^{+}=\frac{1-f}{\sqrt{f(1-f)}}=\sqrt{\frac{1-f}{f}}.
\end{equation}
When consensus is weak, $f$ is small and $\hat A^{+}$ becomes large, causing a small subset of positive pseudo-labeled trajectories to disproportionately influence policy updates. If the voted answer is incorrect, GRPO can rapidly reinforce this spurious signal, driving the policy toward premature convergence on an erroneous solution.

To mitigate the amplification of label noise under group normalization, we adopt a conservative principle: if we cannot credibly identify a correct answer, we abstain from providing positive supervision.
Concretely, we convert majority voting into a selective pseudo-labeling rule with abstention. Given $N$ responses for query $q$, let $\mathcal{A}=\{a_j\}_{j=1}^K$ be the answer distribution with counts $n_j$ and proportions $p_j=n_j/N$. We denote $j^*=\arg\max_{j} p_j$ as the index of the most frequent answer and $p_{(2)}$ as the second-largest proportion. We declare a positive pseudo-label $y^{+}$ only when the answer distribution is sharply concentrated and well-separated. Formally, $y^{+}=a_{j^*}$ if:
\begin{equation}
\label{eq:credible-majority}
p_{j^*} \ge \tau_{\text{pos}} \;\wedge\; \big(p_{j^*}-p_{(2)}\big)> \tau_{\text{marg}},
\end{equation}
otherwise, $y^{+}=\varnothing$.
The threshold $\tau_{\text{pos}}$ prevents positive labeling when the top-ranked answer has insufficient support, while the margin threshold $\tau_{\text{marg}}$ enforces separation from the second-ranked answer, preventing unreliable majorities from being reinforced as supervision signals. When $y^{+}=\varnothing$, we simply avoid positive reinforcement learning, shifting the learning focus to the negative pseudo-labeling described in Section~\ref{sec:pseudo-negative}.

\subsection{Entropy-Gated Negative Pseudo-Labeling}
\label{sec:pseudo-negative}
When the answer distribution is dispersed, reinforcing any single trajectory with a positive label is unreliable. Nevertheless, the model’s responses still contain useful signal: while correct answers are difficult to identify with confidence, incorrect answers can be detected more reliably.
By constructing high-confidence negative labels, we can prune the search space and encourage the model to update toward more plausible regions without forcing a premature collapse \citep{zhu2025surprising}.
\paragraph{Entropy-based uncertainty estimation}
Given responses $\{o_i\}_{i=1}^N$ and the answer distribution
$\mathcal{A}=\{a_j\}_{j=1}^{K}$ with counts $n_j$ and proportions $p_j$, we distinguish between low-frequency but valid answers and incorrect responses by computing an uncertainty measure from the policy. The Shannon entropy of the next-token distribution over the vocabulary $\mathcal{V}$ at step $t$ is:
\begin{equation}
h_{i,t}=-\sum_{v\in\mathcal{V}} \pi_{\text{old}}(v \mid o_{i<t}) \log \pi_{\text{old}}(v \mid o_{i<t}).
\end{equation}
For response $o_i$ with length $T_i$, we derive the trajectory-level uncertainty $\bar h_i$: 
\begin{equation}
\bar h_i \;=\; \frac{1}{T_i}\sum_{t=1}^{T_i} h_{i,t}.
\end{equation}
We then aggregate uncertainty at the answer level:
\begin{equation}
\bar H_j \;=\; \frac{1}{n_j}\sum_{i:\, a_i=a_j} \bar h_i,
\quad
\bar H \;=\; \frac{1}{N}\sum_{i=1}^{N} \bar h_i.
\end{equation}
Intuitively, $\bar H_j$ measures the uncertainty with trajectories leading to answer $a_j$.

\paragraph{Negative pseudo-labeling rule}
We identify answer $a_j$ as a negative pseudo-label that simultaneously (i) has proportion below the low-support threshold $\tau_{\text{neg}}$, and (ii) has generation uncertainty exceeding the query-level average $\bar H$:
\begin{equation} 
\mathcal{N}^- = \left\{ a_j \in \mathcal{A} \mid p_j < \tau_{\text{neg}} \land \bar{H}_j \ge \bar{H} \right\}. 
\end{equation}
The condition $\bar{H}_j \ge \bar{H}$ ensures that we only penalize rare answers with high generation uncertainty \citep{prabhudesai2025maximizing,liang2025comprehensive}, thereby preserving potentially correct low-frequency trajectories until stronger consensus emerges in subsequent iterations.

\subsection{Dynamic Reward Shaping} 
\label{sec:reward-shaping}
In test-time reinforcement learning, assigning fixed rewards to pseudo-labels may amplify noise and destabilize training. To address this, we introduce Dynamic Reward Shaping, which scales reinforcement magnitude based on consensus strength and incorporates uncertainty penalties to guide policy updates.
For each response $o_i$ with answer $a_i$ and proportions $p_i$, the reward is defined as:
\begin{equation}
\begin{aligned}
R_i \;=\; & p(a_i) \mathbb{I}[a_i = y^+] \\
& + \left(p(a_i) - \tau_{\mathrm{neg}}\right) \mathbb{I}[a_i \in \mathcal{N}^-] \\
& - \lambda_H \left(\bar{H}(a_i) - \bar{H}\right).
\end{aligned}
\end{equation}
The first two terms jointly calibrate the reinforcement magnitude based on the strength of the consensus. The final entropy term, weighted by the coefficient $\lambda_H$, gently biases learning toward lower-uncertainty responses. This reward implements a risk-averse strategy that reinforces answers only under credible consensus and eliminates trajectories that are simultaneously rare and uncertain, preventing premature convergence to unreliable solutions.
\section{Experiments}
\subsection{Experimental Setup}
\label{experiment_setup}
\paragraph{Benchmarks and Baselines} 
To evaluate the reasoning capabilities of SCRL, we conduct experiments on six challenging datasets: AIME24 \citep{li2024numinamath}, AIME25 \citep{li2024numinamath}, AMC \citep{li2024numinamath}, MATH-500 \citep{hendrycks2021measuring}, Minerva \citep{team2024qwen2},  and GPQA \citep{rein2024gpqa}. We compare our method against: (1) \textbf{TTRL} \citep{zuo2025ttrl}, which enables the model to self-evolve on unlabeled test data through majority voting. (2) \textbf{COMPASS} \citep{tang2025rewarding}, which introduces a composite reward mechanism that jointly optimizes answer reliability and reasoning quality on unlabeled data. (3) \textbf{ETMR} \citep{liu2025ettrl}, which enhances exploration–exploitation balance through entropy-fork tree rollouts and entropy-based advantage reshaping. (4) \textbf{RESTRAIN} \citep{yu2025restrain}, a self-penalizing framework that penalizes overconfident and low-consistency rollouts to derive useful learning signals without supervision.

\paragraph{Models} To evaluate the generalizability of our method across varying architectures and scales, we utilize a diverse set of open-weight models, including Qwen2.5-3B \citep{yang2025qwen3}, Qwen2.5-Math-7B \citep{yang2025qwen3}, Qwen3-4B \citep{yang2025qwen3}, Llama-3.2-1B-Instruct and Llama-3.1-8B-Instruct \citep{grattafiori2024llama}.

\paragraph{Evaluation Metric} We report pass@1 \citep{guo2025deepseek,chen2021evaluating} as the primary evaluation metric. For each question, we generate $N=16$ responses utilizing a temperature of $0.6$ and a top-$p$ value of $0.95$, with maximum generation length set to 3072 tokens. The pass@1 score is computed as:
$\mathrm{pass@1}=\frac{1}{k}\sum_{i=1}^{k} p_i,$ where $p_i\in\{0,1\}$ indicates the correctness of the $i$-th response.
\paragraph{Hyperparameter Configuration}
We optimize the policy model using the AdamW optimizer with a cosine learning rate schedule peaking at $5 \times 10^{-7}$. During the rollout phase, we generate 64 (or 32) candidate responses for label estimation, employing a temperature of $1.0$ for Qwen2.5-Math and $0.6$ for other models to ensure appropriate exploration. Subsequently, we downsample 32 (or 16) responses per prompt for the training update. The maximum token generation length is set to 3072. For the labeling thresholds, we set $\tau_{\text{pos}}=0.375$, $\tau_{\text{marg}}=0.125$, and $\tau_{\text{neg}}=0.125$ across all experiments. The entropy penalty weight is set to $\lambda_H=0.1$. To accommodate varying dataset sizes and difficulties, we set the number of training episodes to 10 for MATH-500 and Minerva, 30 for AMC, and 80 for AIME, unless otherwise specified. All experiments are conducted on $8\times$ NVIDIA A100 80GB GPUs.

\begin{table*}
  \centering
  \begin{tabular}{lcccccc}
    \hline
    \textbf{Method} & \textbf{AIME25} & \textbf{AMC} & \textbf{MATH-500} & \textbf{Minerva} & \textbf{GPQA} & \textbf{Avg} \\
    \hline
    \multicolumn{7}{c}{\textit{32 candidate responses, 16 training samples}} \\
    \hline
    Qwen2.5-3B      & 1.9   & 23.2  & 50.0   & 21.4 &21.6  & 23.6 \\
    \quad + TTRL \citep{zuo2025ttrl}& 2.6& 39.4& 66.9 & 31.6 &25.0 & 33.1 \\
    \rowcolor[HTML]{D7F6FF}
    \quad + SCRL (Ours)       & 8.4   & 41.5  & 68.2 & 29.5&25.7& 34.7 \\
    \rowcolor[HTML]{D7F6FF}
    \textit{\quad $\Delta$} & \textbf{+5.8} & \textbf{+2.1} & \textbf{+1.3} & \textbf{-2.1}& \textbf{+0.7} & \textbf{+1.6} \\
    \hline
    Qwen2.5-Math-7B & 4.6   &34.0  & 46.5   & 10.1&23.0& 23.6 \\
    \quad + TTRL \citep{zuo2025ttrl} & 16.8  & 65.7& 85.7& 14.5* &25.5&41.6 \\
    \rowcolor[HTML]{D7F6FF}
    \quad + SCRL (Ours)      & 26.9  & 66.9  & 85.6   & 41.6&25.4& 49.3 \\
    \rowcolor[HTML]{D7F6FF}
    \textit{\quad $\Delta$} & \textbf{+10.1} & \textbf{+1.2} & \textbf{-0.1} & \textbf{+27.1}&\textbf{-0.1} & \textbf{+7.7} \\
    \hline
    \multicolumn{7}{c}{\textit{64 candidate responses, 32 training samples}} \\
    \hline
    Qwen2.5-3B      & 1.9   & 23.2  & 50.0   & 21.4&21.6& 23.6 \\
    \quad + TTRL \citep{zuo2025ttrl}& 3.9   & 39.5  & 72.8 & 33.5&23.5& 34.6 \\
    \rowcolor[HTML]{D7F6FF}
    \quad + SCRL (Ours)      & 7.7   & 41.3  & 73.4   & 29.5&24.5& 35.3 \\
    \rowcolor[HTML]{D7F6FF}
\textit{\quad $\Delta$} & \textbf{+3.8} & \textbf{+1.8} & \textbf{+0.6} & \textbf{-4.0} & \textbf{+1.0}& \textbf{+0.7} \\
    \hline
    Qwen2.5-Math-7B & 4.6   & 34.0  & 46.5   & 10.1& 23.0&23.6 \\
    \quad + TTRL \citep{zuo2025ttrl} & 19.0  & 68.1& 83.4& 10.8* &25.5& 41.4 \\
    \rowcolor[HTML]{D7F6FF}
    \quad + SCRL (Ours) & 22.8  & 68.5  & 86.2   & 43.1&26.0 & 49.3 \\
    \rowcolor[HTML]{D7F6FF}
    \textit{\quad $\Delta$} & \textbf{+3.8} & \textbf{+0.4} &\textbf{+2.8} & \textbf{+32.3}&\textbf{+0.5}& \textbf{+7.9} \\
    \hline
  \end{tabular}
  \caption{\label{tab:main_results}
    Main results on various reasoning benchmarks. We report the pass@1 accuracy (\%) across five datasets under two rollout budgets. Results denoted by * represent peak performance before significant degradation during training. \textit{$\Delta$} shows the performance gain over TTRL \citep{zuo2025ttrl}.}
\end{table*}

\begin{table}[h]
\centering
\resizebox{\columnwidth}{!}{%
\begin{tabular}{lccc}
\toprule
\textbf{Method} & \textbf{AIME24} & \textbf{MATH / AMC}$^\star$ & \textbf{Average} \\
\midrule
\multicolumn{4}{c}{\textit{Instruct Models}} \\
\midrule
Llama-3.2-1B$^{\dagger}$            & 1.5  & 24.7 & 13.1 \\
\quad + TTRL$^{\dagger}$             & 6.7  & 27.8 & 17.3 \\
\quad + COMPASS$^{\dagger}$          & 3.5  & 28.7 & 16.1 \\
\rowcolor[HTML]{D7F6FF}
\quad + SCRL                         & 6.7  & 39.7 & 23.2 \\
\rowcolor[HTML]{D7F6FF}
\textit{\quad $\Delta$} & \textbf{+3.2} & \textbf{+11.0} & \textbf{+7.1} \\
\midrule
Llama-3.1-8B$^{\ddagger}$            & 4.6  & 23.3 & 14.0 \\
\quad + TTRL$^{\ddagger}$            & 10.0 & 32.3 & 21.2 \\
\quad + ETMR$^{\ddagger}$            & 16.9 & 35.4 & 26.2 \\
\quad + RESTRAIN$^{\ddagger}$        & 16.7 & 40.0 & 28.4 \\
\rowcolor[HTML]{D7F6FF}
\quad + SCRL                         & 21.9 & 36.1 & 29.0 \\
\rowcolor[HTML]{D7F6FF}
\textit{\quad $\Delta$}& \textbf{+5.2} & \textbf{$-$3.9} & \textbf{+0.6} \\
\bottomrule
\end{tabular}
}
\caption{Pass@1 accuracy (\%) on Llama-3-Instruct models. Column denoted by $^\star$ refers to MATH-500 for Llama-3.2-1B-Instruct and AMC for Llama-3.1-8B-Instruct. Results denoted by $^{\dagger}$ and $^{\ddagger}$ are reported from COMPASS \citep{tang2025rewarding} and RESTRAIN \citep{yu2025restrain}, respectively. \textit{$\Delta$} shows the performance gain over the corresponding baseline.}
\label{tab:main_results2}
\vspace{-1.0\baselineskip}
\end{table}

\begin{figure*}[t]
\vspace{-1.5\baselineskip}
\centering
\begin{subfigure}[b]{0.48\textwidth}
  \centering
  \includegraphics[width=\linewidth]{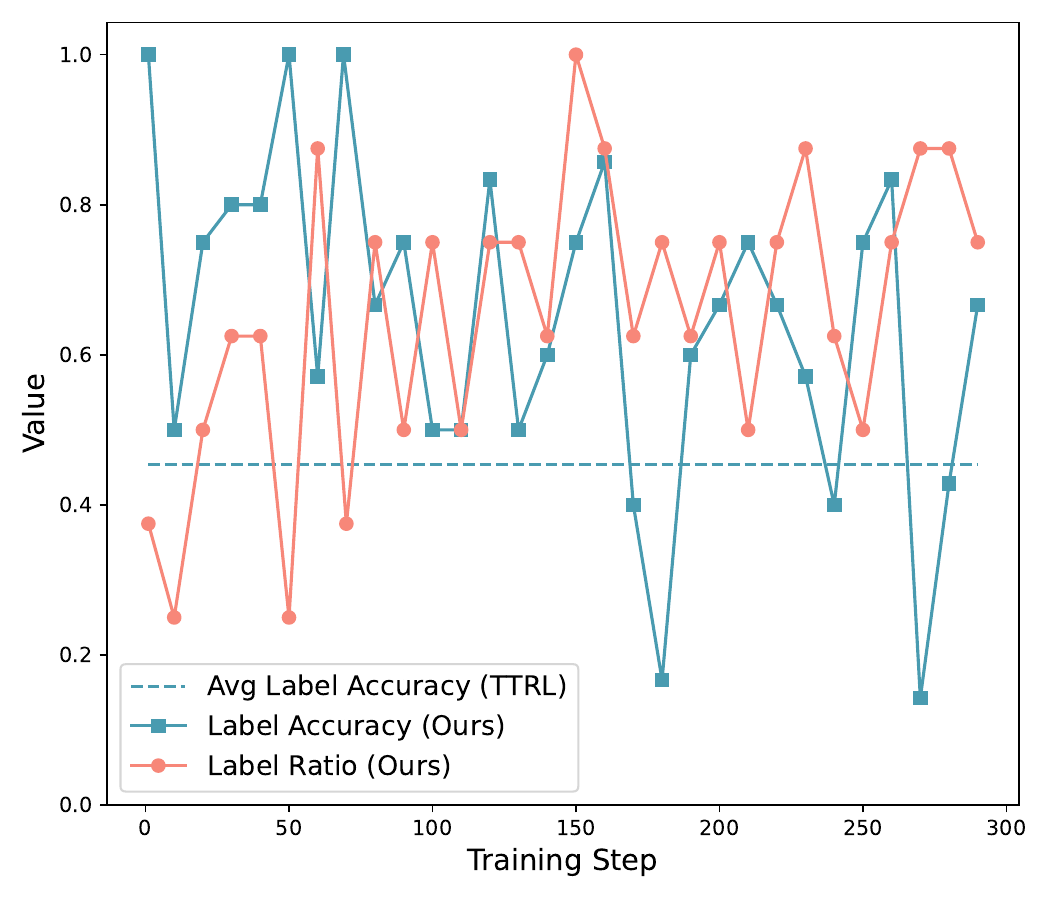}
  \caption{\footnotesize Statistics of positive pseudo-label 
}
  \label{fig:positive_label}
\end{subfigure}
\hfill
\begin{subfigure}[b]{0.48\textwidth}
  \centering
  \includegraphics[width=\linewidth]{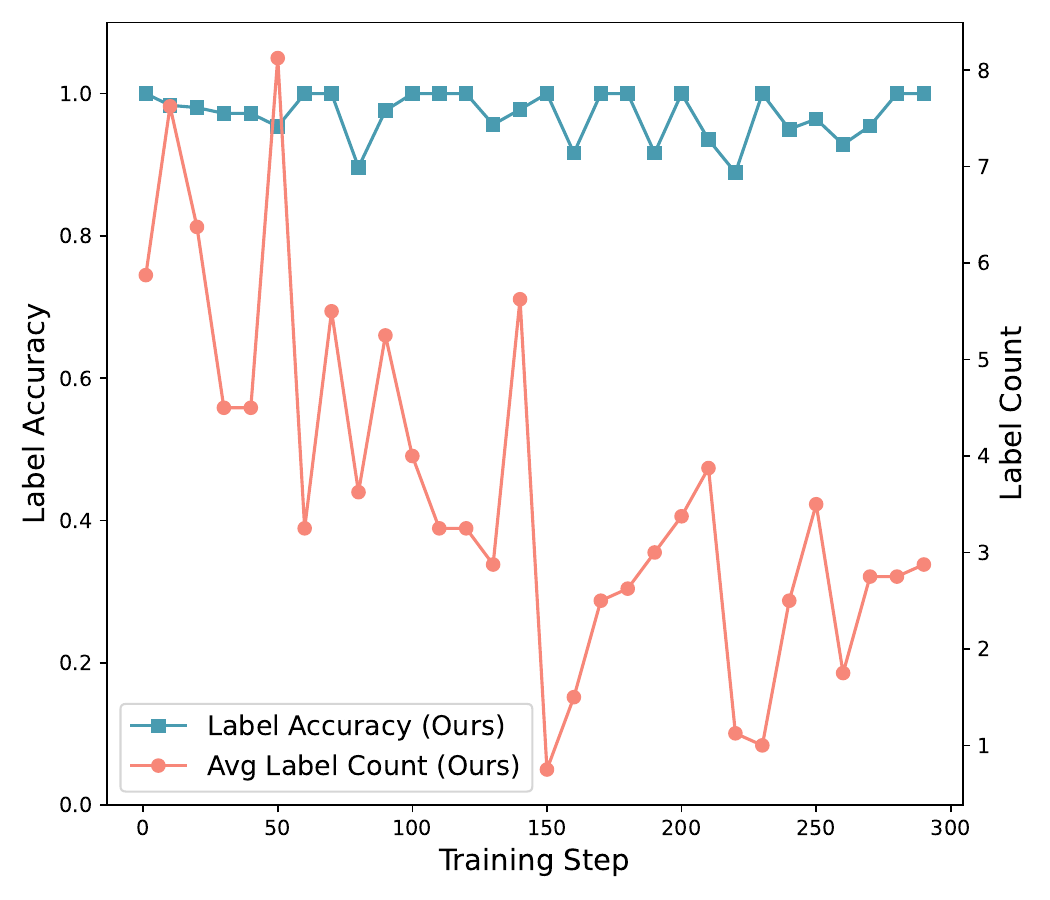}
  \caption{\footnotesize Statistics of negative pseudo-label}
  \label{fig:negative_label}
\end{subfigure}

\caption{Statistics of positive and negative pseudo-label estimation on the AMC dataset using Qwen2.5-3B.}
\label{fig:label_accuracy}
\vspace{-0.5\baselineskip}
\end{figure*}

\subsection{Main Results}

Tables~\ref{tab:main_results} and~\ref{tab:main_results2} present the main results across multiple reasoning benchmarks and models. Our results demonstrate that SCRL consistently achieves substantial improvements over baselines, with particularly pronounced gains on challenging problems and under constrained rollout budgets—precisely the scenarios where weak consensus poses the greatest risk of label noise amplification.
\paragraph{Performance on Vanilla Base Model}
Under the constrained budget setting, SCRL achieves an average improvement of 1.6\% across all benchmarks on Qwen2.5-3B. The most significant improvement is observed on AIME25, where SCRL achieves 8.4\% accuracy compared to TTRL's 2.6\%. This substantial improvement on the most challenging benchmark validates our core hypothesis that SCRL effectively prevents the reinforcement of unreliable majorities when answer distributions are highly dispersed. 
On AMC and MATH-500, SCRL achieves 41.5\% and 68.2\% respectively, compared to TTRL's 39.4\% and 66.9\%, demonstrating consistent gains. 
When the rollout budget is increased to 64 candidate responses with 32 training samples, SCRL maintains an average improvement of 0.7\%, which aligns with theoretical expectations: with more rollouts, majority voting naturally produces more reliable consensus. 
Notably, on AIME25, SCRL maintains a substantial improvement of 3.8\%, demonstrating that for difficult problems, SCRL can remain effective regardless of computational resources.
In addition to mathematics, SCRL consistently improves performance on the GPQA dataset, suggesting that its robustness extends beyond mathematical reasoning to broader general reasoning tasks.
\paragraph{Performance on Math-Specialized Model}
On Qwen2.5-Math-7B, SCRL achieves an average improvement of 7.7\% under the constrained budget. On AIME25, SCRL achieves 26.9\% accuracy compared to TTRL's 16.8\%, an absolute improvement of 10.1\%. 
On AMC and MATH-500, SCRL achieves 66.9\% and 85.6\% respectively, outperforming TTRL on AMC and achieving comparable performance on MATH-500. 
Notably, on Minerva, while TTRL experiences significant training instability and reaches only 14.5\% before performance degradation, SCRL maintains robust training dynamics and achieves 41.6\% accuracy. 
With increased rollout budget of 64 candidate responses and 32 training samples, SCRL sustains its advantage with an average improvement of 7.9\%. On AIME25, SCRL attains 22.8\% compared to TTRL's 19.0\%. 
On AMC and MATH-500, SCRL demonstrates continued improvements. 
Minerva exhibits the most dramatic improvement at 32.3\%, while TTRL still shows training instability even with doubled rollout budget. This indicates the necessity for sophisticated label quality assessment and credible reinforcement mechanisms.
\paragraph{Performance on Instruct Model}
Table~\ref{tab:main_results2} further demonstrates that SCRL generalizes effectively to instruction-tuned models with different architectures.
We evaluate on Llama-3.2-1B-Instruct and Llama-3.1-8B-Instruct using 64 candidate responses and 32 training samples.
On Llama-3.2-1B-Instruct, SCRL achieves an average accuracy of 23.2\%, outperforming both TTRL and COMPASS.
On Llama-3.1-8B-Instruct, SCRL achieves the highest average accuracy of 29.0\%, surpassing all competing baselines including TTRL, ETMR, and RESTRAIN.
The consistency of results across different model families validates that our approach exhibits model-agnostic properties and demonstrates broad applicability across diverse architectures and model scales.

\section{Analysis}

\begin{table}[!htbp]
\centering
\resizebox{0.48\textwidth}{!}{
\begin{tabular}{llll}
\hline
\textbf{Method} & \textbf{AIME25} & \textbf{AMC} \\
\hline
SCRL  &8.4 & 41.5\\
\quad w/o Selective Labeling &3.9$_{\downarrow 4.5}$&41.8$_{\uparrow 0.3}$\\
\quad w/o Negative Labeling &1.2$_{\downarrow 7.2}$&37.7$_{\downarrow 3.8}$\\
\quad w/o Entropy Gate &3.6$_{\downarrow 4.8}$&39.3$_{\downarrow 2.2}$\\
\quad w/o Dynamic Reward &5.1$_{\downarrow 3.3}$&39.8$_{\downarrow 1.7}$\\
\hline
\end{tabular}
}
\caption{Ablation study of SCRL components on Qwen2.5-3B. The table reports pass@1 accuracy (\%).}
\label{tab:ablation}
\end{table}

\subsection{Ablation Study}
To validate the individual contributions of the proposed components in SCRL, we conduct an ablation study on the AIME25 and AMC datasets using Qwen2.5-3B. Table~\ref{tab:ablation} presents the results. Removing the selective positive pseudo-labeling mechanism reduces the positive labeling method to majority voting and causes a substantial performance drop on the AIME25 dataset. This confirms that weak consensus on challenging problems leads to noise amplification when the most frequent answer is reinforced without credibility checks. On the relatively easier AMC dataset, removing the consensus thresholds achieves a slight performance gain, suggesting that when the model is reasonably accurate, stricter thresholds may overly constrain positive supervision.
Furthermore, the substantial degradation when removing negative labeling confirms that negative labeling serves as a complementary signal, pruning the search space and maintaining training stability especially when positive signals are sparse to guide the model.

Removing the entropy gate for negative pseudo-labeling also causes a substantial degradation, indicating that frequency alone is insufficient and uncertainty-aware filtering is necessary to avoid penalizing rare but valid answers. Replacing dynamic reward shaping with hard rewards (+1,0,-1) consistently degrades performance, suggesting that distribution-aware reward magnitudes are important for stabilizing policy updates and calibrating learning to consensus strength.

\begin{table*}[t]
  \centering
  \begin{tabular}{lcccccc}
    \toprule
    \multirow{2}{*}{\textbf{Method}} & \multicolumn{2}{c}{\textbf{AMC}} & \multicolumn{2}{c}{\textbf{MATH-500}} & \multicolumn{2}{c}{\textbf{Minerva}} \\
    \cmidrule(lr){2-3} \cmidrule(lr){4-5} \cmidrule(lr){6-7}
     & Pass@1 & Pass@16 & Pass@1 & Pass@16 & Pass@1 & Pass@16 \\
    \midrule
    Qwen2.5-3B & 23.2&\textbf{67.5} & 50.0 & 86.4 & 21.4 &  \textbf{50.0} \\
    \quad + TTRL \citep{zuo2025ttrl} & 27.3 & 54.2 & 58.5 & 84.8 & 23.4 & 48.9 \\
    \quad + SCRL (Ours)  & \textbf{27.5} & 60.2 & \textbf{65.1} &  \textbf{87.0} &  \textbf{26.3} & 49.3 \\
    \bottomrule
  \end{tabular}
  \caption{Generalization performance of Qwen2.5-3B trained on AIME25. Bold shows the best result.}
  \label{tab:general_results}
\vspace{-0.5\baselineskip}
\end{table*}

\subsection{Statistics of Label Estimation}
To validate the reliability of the supervision signals, we track the dynamics of pseudo-label estimation during training on the AMC dataset using Qwen2.5-3B. As shown in Figure~\ref{fig:positive_label}, our method maintains higher positive label accuracy over TTRL, which confirms that the selective positive labeling mechanism effectively filters out unreliable majorities, particularly in early training steps when the initial policy is weak and answer distributions are highly dispersed. Figure~\ref{fig:negative_label} shows that entropy-gated negative labeling achieves near-perfect accuracy throughout training, which validates our hypothesis that under high uncertainty, incorrect answers can be identified more reliably than correct ones by examining both frequency and generation uncertainty. Meanwhile, the positive label ratio increases with a decrease in negative labels, indicating that the model progressively develops stronger consensus as its reasoning capability improves, and the search space is being effectively pruned through the elimination of implausible trajectories. The complementary dynamics of positive and negative labels demonstrate that SCRL successfully guides the policy toward higher-quality solution regions while maintaining stable supervision signals.
\subsection{Generalization Capabilities}
To evaluate whether SCRL develops transferable reasoning capabilities rather than overfitting to training-specific patterns, we train both SCRL and TTRL on AIME25 and evaluate their generalization performance on three out-of-distribution datasets. We use Qwen2.5-3B as the backbone and report both pass@1 and pass@16 results. As shown in Table~\ref{tab:general_results}, SCRL consistently improves pass@1 accuracy over TTRL across all three benchmarks. Notably, while TTRL's pass@16 performance degrades substantially from the base model, SCRL maintains more stable pass@16 scores. This indicates that majority-voting mechanism narrows the solution space during training, whereas SCRL's conservative labeling strategy preserves exploration capacity while improving answer quality. These results confirm that selective positive labeling and entropy-gated negative pruning generalize robustly without overfitting to the training dataset.
\subsection{Analysis of Labeling Thresholds}
To investigate the sensitivity of SCRL to the labeling threshold hyperparameters, we conduct an analysis of $\tau_{\text{pos}}$ and $\tau_{\text{neg}}$ using Qwen2.5-Math-7B with 64 candidate responses and 32 training samples. Table~\ref{tab:hyperment} presents the results across different threshold configurations.
Increasing $\tau_{\text{pos}}$ from 0.25 to 0.375 consistently improves performance on AIME25. This substantial gain on the most challenging benchmark validates our hypothesis that stricter consensus criteria are crucial for difficult problems where answer distributions are highly dispersed. The more conservative threshold effectively filters out unreliable majorities that would introduce label noise. 
The impact of $\tau_{\text{neg}}$ exhibits a complementary pattern. On AMC and MATH-500, increasing $\tau_{\text{neg}}$ from 0.0625 to 0.125 generally improves performance, indicating that keeping a moderately larger set of negatives can strengthen the contrastive learning signal and enhance exploration of alternative solution strategies.
The performance gaps across different configurations become more pronounced on harder problems, reinforcing that SCRL's conservative labeling strategy is particularly beneficial in scenarios where weak consensus poses the risk of label noise amplification. See Appendix~\ref{sec:margin_analysis} for analysis of $\tau_{\text{marg}}$ and $\lambda_H$.

\begin{table}[!htbp]
\centering
\begin{tabular}{ll|ccc}
\hline
$\tau_{\text{pos}}$ &$\tau_{\text{neg}}$ &\textbf{AIME25}& \textbf{AMC} &\textbf{MATH} \\
\hline
0.25 & 0.0625&19.2& 64.5&85.5\\
0.25 & 0.125&19.0&66.6&85.4\\
0.375 & 0.0625&22.1&66.3&85.0\\
0.375 & 0.125&22.8&67.5&86.2\\
\hline
\end{tabular}
\caption{Parameter analysis of $\tau_{\text{pos}}$ and $\tau_{\text{neg}}$ on Qwen2.5-Math-7B. The table reports pass@1 accuracy (\%).}
\label{tab:hyperment}
\vspace{-0.5\baselineskip}
\end{table}

\begin{table*}[!htbp]
\centering
\begin{tabular}{lcccc}
\hline
\textbf{Type} & \textbf{AIME24} & \textbf{AMC}&\textbf{MATH-500}&\textbf{Minerva} \\
\hline
Correct Trajectories &0.208&0.225&0.245&0.258\\
Incorrect Trajectories  &0.288&0.279&0.272&0.316\\
\rowcolor[HTML]{D7F6FF}
Overall Average&0.286&0.266&0.259&0.303\\
\hline
\end{tabular}
\caption{Average trajectory-level entropy for correct and incorrect answers across mathematical reasoning benchmarks. Results are computed using Qwen2.5-3B during the initial rollout phase.}
\label{tab:entropy}
\vspace{-0.5\baselineskip}
\end{table*}

\subsection{Entropy Analysis}
To validate the effectiveness of our entropy-gated negative labeling mechanism, we analyze the relationship between trajectory-level entropy and answer correctness. As shown in Table~\ref{tab:entropy}, we compute the average trajectory entropy $\bar{h}_i$ for correct and incorrect trajectories across multiple benchmarks using the Qwen2.5-3B model. The results reveal a consistent and substantial gap between correct and incorrect trajectories across all benchmarks, which validates our hypothesis that generation uncertainty serves as a robust signal for distinguishing implausible solutions. This empirical evidence supports the entropy-gated mechanism, which leverages this uncertainty differential to construct high-confidence negative labels while preserving potentially correct low-frequency trajectories.

\subsection{Performance on Large Reasoning Models}
To evaluate SCRL's applicability to long chain-of-thought reasoning models \citep{wei2023chainofthoughtpromptingelicitsreasoning}, we conduct experiments using Qwen3-4B with thinking mode enabled on the AIME25 dataset.
As shown in Table~\ref{tab:lrm}, SCRL consistently outperforms TTRL across both maximum generation length settings, achieving +2.7\% and +0.9\% improvements, respectively.
These results confirm that selective positive labeling and entropy-gated negative pruning remain effective on long-CoT reasoning models with substantially extended generation lengths.

\begin{table}[!htbp]
\centering
\begin{tabular}{lcc}
\hline
\multirow{2}{*}{\textbf{Method}} & \multicolumn{2}{c}{\textbf{Max Generation Length}} \\
\cmidrule(lr){2-3}
& \textbf{10,240} & \textbf{15,360} \\
\hline
Qwen3-4B&38.9&51.9\\
\quad + TTRL &50.9&59.3\\
\rowcolor[HTML]{D7F6FF}
\quad + SCRL &53.6&60.2\\
\rowcolor[HTML]{D7F6FF}
\textit{\quad $\Delta$}&\textbf{+2.7}&\textbf{+0.9}\\
\hline
\end{tabular}
\caption{Pass@1 accuracy (\%) on large reasoning models using Qwen3-4B under maximum generation length of 10k and 15k tokens. \textit{$\Delta$} shows the performance gain over TTRL \citep{zuo2025ttrl}.}
\label{tab:lrm}
\vspace{-0.5\baselineskip}
\end{table}

\subsection{Failure Analysis}
SCRL underperforms TTRL on the Minerva dataset with Qwen2.5-3B, contrasting with consistent improvements on other benchmarks. Our analysis suggests this discrepancy arises from the interaction between the smaller model's limited domain-specific knowledge and the strictness of our selective labeling mechanism.

The Minerva dataset consists of physics-based mathematical problems requiring specialized domain knowledge beyond pure mathematical reasoning. The 3B model, lacking sufficient capacity to store such extensive domain knowledge, produces highly dispersed answer distributions. This dispersion is further exacerbated by the nature of physics-based mathematical problems in Minerva, which involve variations in numerical precision and equivalent formulations, fragmenting the correct solution across similar but non-identical answers. Under these conditions, SCRL's selective positive labeling becomes over-conservative, frequently abstaining when strict thresholds are unmet, starving the model of positive supervision. 

\section{Conclusion}
In this work, we propose SCRL, a novel framework for test-time reinforcement learning to mitigate label noise amplification in unsupervised settings. SCRL integrates Selective Positive Pseudo-Labeling, which enforces strict consensus criteria to prevent reinforcing unreliable majorities under weak consensus, with Entropy-Gated Negative Pseudo-Labeling, the first negative supervision mechanism in test-time reinforcement learning, which reliably prunes implausible solutions by isolating answers with both low frequency and high uncertainty. This dual mechanism ensures that the model reinforces high-confidence trajectories while reliably identifying and penalizing incorrect answers, without discarding rare but potentially correct trajectories. Empirical results confirm that SCRL significantly outperforms baselines, delivering robust generalization and training stability.

\section*{Limitations}
While SCRL demonstrates substantial empirical improvements across multiple benchmarks, several limitations remain to be addressed in future work. First, SCRL lacks theoretical analysis for the effectiveness of selective positive pseudo-labeling and entropy-gated negative pseudo-labeling. Although we provide empirical validation through label accuracy statistics and ablation studies, we have not established theoretical guarantees on conditions where our mechanisms provably outperform standard majority voting. Second, the comparison with concurrent methods is constrained by reproducibility challenges. Most related works in the TTRL paradigm have not released code and critical hyperparameter settings cannot be precisely aligned, potentially limiting the scope of our baseline analysis.

\section*{Acknowledgements}
This research is supported by the National Natural Science Foundation of China under Grants~(62276256, U2441251).
We thank Yongcan Yu at NLPR for early discussions and feedback to this project. 
Besides, we also extend our sincere thanks to the anonymous reviewers for their constructive suggestions.
\nocite{article}
\bibliography{main}

@inproceedings{zuo2025ttrl,
  title={Ttrl: Test-time reinforcement learning},
  author={Zuo, Yuxin and Zhang, Kaiyan and Sheng, Li and Qu, Shang and Cui, Ganqu and Zhu, Xuekai and Li, Haozhan and Zhang, Yuchen and Long, Xinwei and Hua, Ermo and others},
  booktitle={Proc.\ NeurIPS},
  year={2025}
}

@inproceedings{wei2025unsupervised,
  title={Unsupervised Post-Training for Multi-Modal LLM Reasoning via GRPO},
  author={Wei, Lai and Li, Yuting and Wang, Chen and Wang, Yue and Kong, Linghe and Huang, Weiran and Sun, Lichao},
  booktitle={Proc.\ NeurIPS},
  year={2025}
}

@article{yu2025restrain,
  title={RESTRAIN: From Spurious Votes to Signals--Self-Driven RL with Self-Penalization},
  author={Yu, Zhaoning and Su, Will and Tao, Leitian and Wang, Haozhu and Singh, Aashu and Yu, Hanchao and Wang, Jianyu and Gao, Hongyang and Yuan, Weizhe and Weston, Jason and others},
  journal={arXiv preprint arXiv:2510.02172},
  year={2025}
}

@article{liu2025ettrl,
  title={Ettrl: Balancing exploration and exploitation in llm test-time reinforcement learning via entropy mechanism},
  author={Liu, Jia and He, ChangYi and Lin, YingQiao and Yang, MingMin and Shen, FeiYang and Liu, ShaoGuo},
  journal={arXiv preprint arXiv:2508.11356},
  year={2025}
}

@article{wang2025self,
  title={Self-Harmony: Learning to Harmonize Self-Supervision and Self-Play in Test-Time Reinforcement Learning},
  author={Wang, Ru and Huang, Wei and Cao, Qi and Iwasawa, Yusuke and Matsuo, Yutaka and Guo, Jiaxian},
  journal={arXiv preprint arXiv:2511.01191},
  year={2025}
}

@article{wu2025spine,
  title={SPINE: Token-Selective Test-Time Reinforcement Learning with Entropy-Band Regularization},
  author={Wu, Jianghao and George, Yasmeen and Ye, Jin and Wu, Yicheng and Schmidt, Daniel F and Cai, Jianfei},
  journal={arXiv preprint arXiv:2511.17938},
  year={2025}
}

@article{prabhudesai2025maximizing,
  title={Maximizing Confidence Alone Improves Reasoning},
  author={Prabhudesai, Mihir and Chen, Lili and Ippoliti, Alex and Fragkiadaki, Katerina and Liu, Hao and Pathak, Deepak},
  journal={arXiv preprint arXiv:2505.22660},
  year={2025}
}

@inproceedings{ouyang2022training,
  title={Training language models to follow instructions with human feedback},
  author={Ouyang, Long and Wu, Jeffrey and Jiang, Xu and Almeida, Diogo and Wainwright, Carroll and Mishkin, Pamela and Zhang, Chong and Agarwal, Sandhini and Slama, Katarina and Ray, Alex and others},
  booktitle={Proc.\ NeurIPS},
  year={2022}
}

@article{schulman2017proximal,
  title={Proximal policy optimization algorithms},
  author={Schulman, John and Wolski, Filip and Dhariwal, Prafulla and Radford, Alec and Klimov, Oleg},
  journal={arXiv preprint arXiv:1707.06347},
  year={2017}
}

@inproceedings{rafailov2023direct,
  title={Direct preference optimization: Your language model is secretly a reward model},
  author={Rafailov, Rafael and Sharma, Archit and Mitchell, Eric and Manning, Christopher D and Ermon, Stefano and Finn, Chelsea},
  booktitle={Proc.\ NeurIPS},
  year={2023}
}

@inproceedings{meng2024simpo,
  title={Simpo: Simple preference optimization with a reference-free reward},
  author={Meng, Yu and Xia, Mengzhou and Chen, Danqi},
  booktitle={Proc.\ NeurIPS},
  year={2024}
}

@article{shao2024deepseekmath,
  title={Deepseekmath: Pushing the limits of mathematical reasoning in open language models},
  author={Shao, Zhihong and Wang, Peiyi and Zhu, Qihao and Xu, Runxin and Song, Junxiao and Bi, Xiao and Zhang, Haowei and Zhang, Mingchuan and Li, YK and Wu, Yang and others},
  journal={arXiv preprint arXiv:2402.03300},
  year={2024}
}

@article{lambert2024tulu,
  title={Tulu 3: Pushing frontiers in open language model post-training},
  author={Lambert, Nathan and Morrison, Jacob and Pyatkin, Valentina and Huang, Shengyi and Ivison, Hamish and Brahman, Faeze and Miranda, Lester James V and Liu, Alisa and Dziri, Nouha and Lyu, Shane and others},
  journal={arXiv preprint arXiv:2411.15124},
  year={2024}
}

@article{yang2025qwen3,
  title={Qwen3 technical report},
  author={Yang, An and Li, Anfeng and Yang, Baosong and Zhang, Beichen and Hui, Binyuan and Zheng, Bo and Yu, Bowen and Gao, Chang and Huang, Chengen and Lv, Chenxu and others},
  journal={arXiv preprint arXiv:2505.09388},
  year={2025}
}

@article{guo2025deepseek,
  title={Deepseek-r1: Incentivizing reasoning capability in llms via reinforcement learning},
  author={Guo, Daya and Yang, Dejian and Zhang, Haowei and Song, Junxiao and Zhang, Ruoyu and Xu, Runxin and Zhu, Qihao and Ma, Shirong and Wang, Peiyi and Bi, Xiao and others},
  journal={arXiv preprint arXiv:2501.12948},
  year={2025}
}

@article{zhao2025learning,
  title={Learning to reason without external rewards},
  author={Zhao, Xuandong and Kang, Zhewei and Feng, Aosong and Levine, Sergey and Song, Dawn},
  journal={arXiv preprint arXiv:2505.19590},
  year={2025}
}

@article{zhang2025right,
  title={Right question is already half the answer: Fully unsupervised llm reasoning incentivization},
  author={Zhang, Qingyang and Wu, Haitao and Zhang, Changqing and Zhao, Peilin and Bian, Yatao},
  journal={arXiv preprint arXiv:2504.05812},
  year={2025}
}

@article{zhang2025consistent,
  title={Consistent Paths Lead to Truth: Self-Rewarding Reinforcement Learning for LLM Reasoning},
  author={Zhang, Kongcheng and Yao, Qi and Liu, Shunyu and Wang, Yingjie and Lai, Baisheng and Ye, Jieping and Song, Mingli and Tao, Dacheng},
  journal={arXiv preprint arXiv:2506.08745},
  year={2025}
}

@article{li2024numinamath,
  title={Numinamath: The largest public dataset in ai4maths with 860k pairs of competition math problems and solutions},
  author={Li, Jia and Beeching, Edward and Tunstall, Lewis and Lipkin, Ben and Soletskyi, Roman and Huang, Shengyi and Rasul, Kashif and Yu, Longhui and Jiang, Albert Q and Shen, Ziju and others},
  journal={Hugging Face repository},
  volume={13},
  number={9},
  pages={9},
  year={2024}
}

@inproceedings{hendrycks2021measuring,
  title={Measuring mathematical problem solving with the math dataset},
  author={Hendrycks, Dan and Burns, Collin and Kadavath, Saurav and Arora, Akul and Basart, Steven and Tang, Eric and Song, Dawn and Steinhardt, Jacob},
  booktitle={Proc.\ NeurIPS},
  year={2021}
}

@article{team2024qwen2,
  title={Qwen2 technical report},
  author={Team, Qwen and others},
  journal={arXiv preprint arXiv:2407.10671},
  volume={2},
  number={3},
  year={2024}
}

@article{chen2021evaluating,
  title={Evaluating large language models trained on code},
  author={Chen, Mark},
  journal={arXiv preprint arXiv:2107.03374},
  year={2021}
}

@article{grattafiori2024llama,
  title={The llama 3 herd of models},
  author={Grattafiori, Aaron and Dubey, Abhimanyu and Jauhri, Abhinav and Pandey, Abhinav and Kadian, Abhishek and Al-Dahle, Ahmad and Letman, Aiesha and Mathur, Akhil and Schelten, Alan and Vaughan, Alex and others},
  journal={arXiv preprint arXiv:2407.21783},
  year={2024}
}

@inproceedings{zhu2025surprising,
  title={The surprising effectiveness of negative reinforcement in LLM reasoning},
  author={Zhu, Xinyu and Xia, Mengzhou and Wei, Zhepei and Chen, Wei-Lin and Chen, Danqi and Meng, Yu},
  booktitle={Proc.\ NeurIPS},
  year={2025}
}

@article{jaech2024openai,
  title={Openai o1 system card},
  author={Jaech, Aaron and Kalai, Adam and Lerer, Adam and Richardson, Adam and El-Kishky, Ahmed and Low, Aiden and Helyar, Alec and Madry, Aleksander and Beutel, Alex and Carney, Alex and others},
  journal={arXiv preprint arXiv:2412.16720},
  year={2024}
}

@article{gao2024designing,
  title={On designing effective rl reward at training time for llm reasoning},
  author={Gao, Jiaxuan and Xu, Shusheng and Ye, Wenjie and Liu, Weilin and He, Chuyi and Fu, Wei and Mei, Zhiyu and Wang, Guangju and Wu, Yi},
  journal={arXiv preprint arXiv:2410.15115},
  year={2024}
}

@inproceedings{setlur2024rl,
  title={Rl on incorrect synthetic data scales the efficiency of llm math reasoning by eight-fold},
  author={Setlur, Amrith and Garg, Saurabh and Geng, Xinyang and Garg, Naman and Smith, Virginia and Kumar, Aviral},
  booktitle={Proc.\ NeurIPS},
  year={2024}
}

@article{wang2024enhancing,
  title={Enhancing code llms with reinforcement learning in code generation: A survey},
  author={Wang, Junqiao and Zhang, Zeng and He, Yangfan and Zhang, Zihao and Song, Xinyuan and Song, Yuyang and Shi, Tianyu and Li, Yuchen and Xu, Hengyuan and Wu, Kunyu and others},
  journal={arXiv preprint arXiv:2412.20367},
  year={2024}
}

@article{shafayat2025can,
  title={Can Large Reasoning Models Self-Train?},
  author={Shafayat, Sheikh and Tajwar, Fahim and Salakhutdinov, Ruslan and Schneider, Jeff and Zanette, Andrea},
  journal={arXiv preprint arXiv:2505.21444},
  year={2025}
}

@inproceedings{prasad2024self,
  title={Self-consistency preference optimization},
  author={Prasad, Archiki and Yuan, Weizhe and Pang, Richard Yuanzhe and Xu, Jing and Fazel-Zarandi, Maryam and Bansal, Mohit and Sukhbaatar, Sainbayar and Weston, Jason and Yu, Jane},
    booktitle={Proc.\ ICML},
  year={2025}
}

@article{liu2025understanding,
  title={Understanding r1-zero-like training: A critical perspective},
  author={Liu, Zichen and Chen, Changyu and Li, Wenjun and Qi, Penghui and Pang, Tianyu and Du, Chao and Lee, Wee Sun and Lin, Min},
  journal={arXiv preprint arXiv:2503.20783},
  year={2025}
}

@inproceedings{stahlberg2022uncertainty,
  title={Uncertainty determines the adequacy of the mode and the tractability of decoding in sequence-to-sequence models},
  author={Stahlberg, Felix and Kulikov, Ilia and Kumar, Shankar},
  booktitle={Proc.\ ACL},
  year={2022}
}

@article{shi2025heimdall,
  title={Heimdall: test-time scaling on the generative verification},
  author={Shi, Wenlei and Jin, Xing},
  journal={arXiv preprint arXiv:2504.10337},
  year={2025}
}

@inproceedings{huang2024mirror,
  title={Mirror-consistency: Harnessing inconsistency in majority voting},
  author={Huang, Siyuan and Ma, Zhiyuan and Du, Jintao and Meng, Changhua and Wang, Weiqiang and Lin, Zhouhan},
  booktitle={Proc.\ EMNLP},
  year={2024}
}

@article{tang2025rewarding,
  title={Rewarding the Journey, Not Just the Destination: A Composite Path and Answer Self-Scoring Reward Mechanism for Test-Time Reinforcement Learning},
  author={Tang, Chenwei and Xing, Jingyu and Long, Lin and Liu, Xinyu and Xiong, Deng and Ju, Wei and Huang, Shudong and Lv, Jiancheng and Qiao, Ziyue},
  journal={arXiv preprint arXiv:2510.17923},
  year={2025}
}

@article{yang2025spell,
  title={Spell: Self-play reinforcement learning for evolving long-context language models},
  author={Yang, Ziyi and Shen, Weizhou and Chen, Ruijun and Li, Chenliang and Wan, Fanqi and Yan, Ming and Quan, Xiaojun and Huang, Fei},
  journal={arXiv preprint arXiv:2509.23863},
  year={2025}
}

@article{tan2025diagnosing,
  title={Diagnosing and Mitigating System Bias in Self-Rewarding RL},
  author={Tan, Chuyi and Yuan, Peiwen and Wang, Xinglin and Li, Yiwei and Feng, Shaoxiong and Zhang, Yueqi and Shi, Jiayi and Zhang, Ji and Pan, Boyuan and Hu, Yao and others},
  journal={arXiv preprint arXiv:2510.08977},
  year={2025}
}

@article{feng2025don,
  title={Don't Waste Mistakes: Leveraging Negative RL-Groups via Confidence Reweighting},
  author={Feng, Yunzhen and Jain, Parag and Hartshorn, Anthony and Duan, Yaqi and Kempe, Julia},
  journal={arXiv preprint arXiv:2510.08696},
  year={2025}
}

@article{wen2025self,
  title={Self-Evolving Vision-Language Models for Image Quality Assessment via Voting and Ranking},
  author={Wen, Wen and Zhi, Tianwu and Fan, Kanglong and Li, Yang and Peng, Xinge and Zhang, Yabin and Liao, Yiting and Li, Junlin and Zhang, Li},
  journal={arXiv preprint arXiv:2509.25787},
  year={2025}
}

@article{zhou2025evolving,
  title={Evolving language models without labels: Majority drives selection, novelty promotes variation},
  author={Zhou, Yujun and Liang, Zhenwen and Liu, Haolin and Yu, Wenhao and Panaganti, Kishan and Song, Linfeng and Yu, Dian and Zhang, Xiangliang and Mi, Haitao and Yu, Dong},
  journal={arXiv preprint arXiv:2509.15194},
  year={2025}
}

@article{jayalath2025compute,
  title={Compute as teacher: Turning inference compute into reference-free supervision},
  author={Jayalath, Dulhan and Goel, Shashwat and Foster, Thomas and Jain, Parag and Gururangan, Suchin and Zhang, Cheng and Goyal, Anirudh and Schelten, Alan},
  journal={arXiv preprint arXiv:2509.14234},
  year={2025}
}

@article{yuan2025wisdom,
  title={Wisdom of the Crowd: Reinforcement Learning from Coevolutionary Collective Feedback},
  author={Yuan, Wenzhen and Tang, Shengji and Lin, Weihao and Ruan, Jiacheng and Cui, Ganqu and Zhang, Bo and Chen, Tao and Liu, Ting and Fu, Yuzhuo and Ye, Peng and others},
  journal={arXiv preprint arXiv:2508.12338},
  year={2025}
}

@article{yu2025dapo,
  title={Dapo: An open-source llm reinforcement learning system at scale},
  author={Yu, Qiying and Zhang, Zheng and Zhu, Ruofei and Yuan, Yufeng and Zuo, Xiaochen and Yue, Yu and Dai, Weinan and Fan, Tiantian and Liu, Gaohong and Liu, Lingjun and others},
  journal={arXiv preprint arXiv:2503.14476},
  year={2025}
}

@inproceedings{wang2025reinforcement,
  title={Reinforcement learning for reasoning in large language models with one training example},
  author={Wang, Yiping and Yang, Qing and Zeng, Zhiyuan and Ren, Liliang and Liu, Liyuan and Peng, Baolin and Cheng, Hao and He, Xuehai and Wang, Kuan and Gao, Jianfeng and others},
  booktitle={Proc.\ NeurIPS},
  year={2025}
}

@article{cui2025process,
  title={Process reinforcement through implicit rewards},
  author={Cui, Ganqu and Yuan, Lifan and Wang, Zefan and Wang, Hanbin and Zhang, Yuchen and Chen, Jiacheng and Li, Wendi and He, Bingxiang and Fan, Yuchen and Yu, Tianyu and others},
  journal={arXiv preprint arXiv:2502.01456},
  year={2025}
}

@article{zhao2025absolute,
  title={Absolute zero: Reinforced self-play reasoning with zero data},
  author={Zhao, Andrew and Wu, Yiran and Yue, Yang and Wu, Tong and Xu, Quentin and Lin, Matthieu and Wang, Shenzhi and Wu, Qingyun and Zheng, Zilong and Huang, Gao},
  journal={arXiv preprint arXiv:2505.03335},
  year={2025}
}

@inproceedings{sheng2025hybridflow,
  title={Hybridflow: A flexible and efficient rlhf framework},
  author={Sheng, Guangming and Zhang, Chi and Ye, Zilingfeng and Wu, Xibin and Zhang, Wang and Zhang, Ru and Peng, Yanghua and Lin, Haibin and Wu, Chuan},
  booktitle={Proceedings of the Twentieth European Conference on Computer Systems},
  pages={1279--1297},
  year={2025}
}

@inproceedings{rein2024gpqa,
  title={Gpqa: A graduate-level google-proof q\&a benchmark},
  author={Rein, David and Hou, Betty Li and Stickland, Asa Cooper and Petty, Jackson and Pang, Richard Yuanzhe and Dirani, Julien and Michael, Julian and Bowman, Samuel R},
  booktitle={First conference on language modeling},
  year={2024}
}

@article{yan2025mission,
  title={Mission Impossible: Feedback-Guided Dynamic Interactive Planning for Improving Reasoning on LLMs},
  author={Yan, Dong and Wu, Gaochen and Zhou, Bowen},
  journal={arXiv preprint arXiv:2510.05577},
  year={2025}
}

@misc{wei2023chainofthoughtpromptingelicitsreasoning,
      title={Chain-of-Thought Prompting Elicits Reasoning in Large Language Models}, 
      author={Jason Wei and Xuezhi Wang and Dale Schuurmans and Maarten Bosma and Brian Ichter and Fei Xia and Ed Chi and Quoc Le and Denny Zhou},
      year={2023},
      eprint={2201.11903},
      archivePrefix={arXiv},
      primaryClass={cs.CL},
      url={https://arxiv.org/abs/2201.11903}, 
}

@inproceedings{lu2025uni,
  title={Uni-layout: Integrating human feedback in unified layout generation and evaluation},
  author={Lu, Shuo and Chen, Yanyin and Feng, Wei and Fan, Jiahao and Li, Fengheng and Zhang, Zheng and Lv, Jingjing and Shen, Junjie and Law, Ching and Liang, Jian},
  booktitle={Proceedings of the 33rd ACM International Conference on Multimedia},
  pages={7709--7718},
  year={2025}
}

@article{lu2025deepresearch,
  title={DeepResearch-Slice: Bridging the Retrieval-Utilization Gap via Explicit Text Slicing},
  author={Lu, Shuo and Xu, Yinuo and Cheng, Jianjie and He, Lingxiao and Wang, Meng and Liang, Jian},
  journal={arXiv preprint arXiv:2601.03261},
  year={2025}
}

@article{liang2025comprehensive,
  title={A comprehensive survey on test-time adaptation under distribution shifts},
  author={Liang, Jian and He, Ran and Tan, Tieniu},
  journal={International Journal of Computer Vision},
  volume={133},
  number={1},
  pages={31--64},
  year={2025},
  publisher={Springer}
}

@inproceedings{yu2026understanding,
  title={Understanding and Mitigating Spurious Signal Amplification in Test-Time Reinforcement Learning for Math Reasoning
},
  author={Yu, Yongcan and He, Lingxiao and Liang, Jian and Guo, Kuangpu and Wang, Meng and Xie, Qianlong and Wang, Xingxing and He, Ran},
  booktitle={Findings of the Association for Computational Linguistics: ACL 2026},
  year={2026}
}

@article{article,
author = {Duan, Junxian and Liu, Siyu and Hao, Yiming and Huang, Huaibo and He, Ran},
year = {2025},
title = {Dual Frequency-Guided Spatiotemporal Feature Learning for Face Forgery Detection},
journal = {IEEE Transactions on Biometrics, Behavior, and Identity Science},
doi = {10.1109/TBIOM.2025.3646181}
}

@misc{wang2025comprehensivesurveytrustworthinessreasoning,
      title={A Comprehensive Survey on Trustworthiness in Reasoning with Large Language Models}, 
      author={Yanbo Wang and Yongcan Yu and Jian Liang and Ran He},
      year={2025},
      eprint={2509.03871},
      archivePrefix={arXiv},
      primaryClass={cs.CL},
      url={https://arxiv.org/abs/2509.03871}, 
}

\appendix

\section{Implementation Details}

\subsection{Prompt Design}
Consistent with TTRL \citep{zuo2025ttrl}, we adopt the standard chat templates corresponding to each model architecture.
For \textbf{Qwen2.5-3B}, we employ the following prompt template:
\begin{tcolorbox}[colback=white, colframe=black, boxrule=0.5mm, arc=4mm, width=0.48\textwidth]
system\\
You are a helpful assistant.\\
user\\
\{question\} Let's think step by step and output the final answer within \verb|\\boxed{}|.\\
assistant
\end{tcolorbox}

For \textbf{Qwen2.5-Math-7B}, we employ the following prompt template:
\begin{tcolorbox}[colback=white, colframe=black, boxrule=0.5mm, arc=4mm, width=0.48\textwidth]
system\\
Please reason step by step, and put your final answer within \verb|\\boxed{}|.\\
user\\
\{question\}\\
assistant
\end{tcolorbox}

\begin{table*}[!htbp]
\vspace{-1.0\baselineskip}
\centering
\begin{tabular}{lcc}
\hline
\textbf{Hyperparameter} & \textbf{Qwen2.5-3B} & \textbf{Qwen2.5-Math-7B} \\
\hline
Max Prompt Length  &512&512\\
Max Response Length  &3072&3072\\
Train Batch Size &8&8\\
PPO Mini Batch Size &1&1\\
PPO Micro Batch Size per GPU &2&2\\
N GPUs per Node&4&8\\
Learning Rate&$5 \times 10^{-7}$&$5 \times 10^{-7}$\\
LR Warmup Steps Ratio &0.03&0.03\\
Warmup Style &Cosine&Cosine\\
Temperature &0.6&1.0\\
Optimizer &AdamW&AdamW\\
\hline
\end{tabular}
\caption{Training hyperparameter configuration for Qwen2.5-3B and Qwen2.5-Math-7B based on the verl \citep{sheng2025hybridflow} framework.}
\label{tab:parameter}
\end{table*}
\begin{figure*}[h]
  \centering
  \begin{subfigure}[t]{0.32\textwidth}
    \centering
    \includegraphics[width=\linewidth]{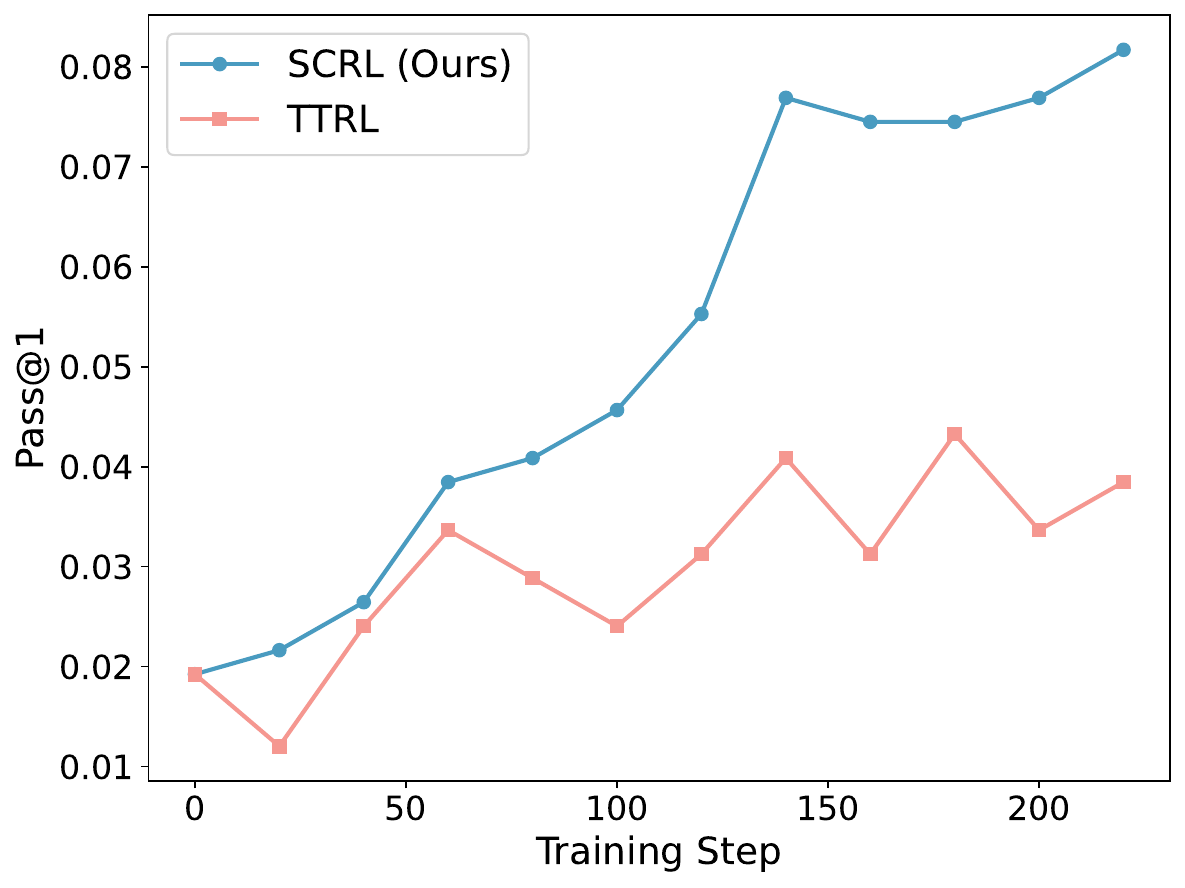}
    \caption{AIME25}
    \label{fig:a}
  \end{subfigure}\hfill
  \begin{subfigure}[t]{0.32\textwidth}
    \centering
    \includegraphics[width=\linewidth]{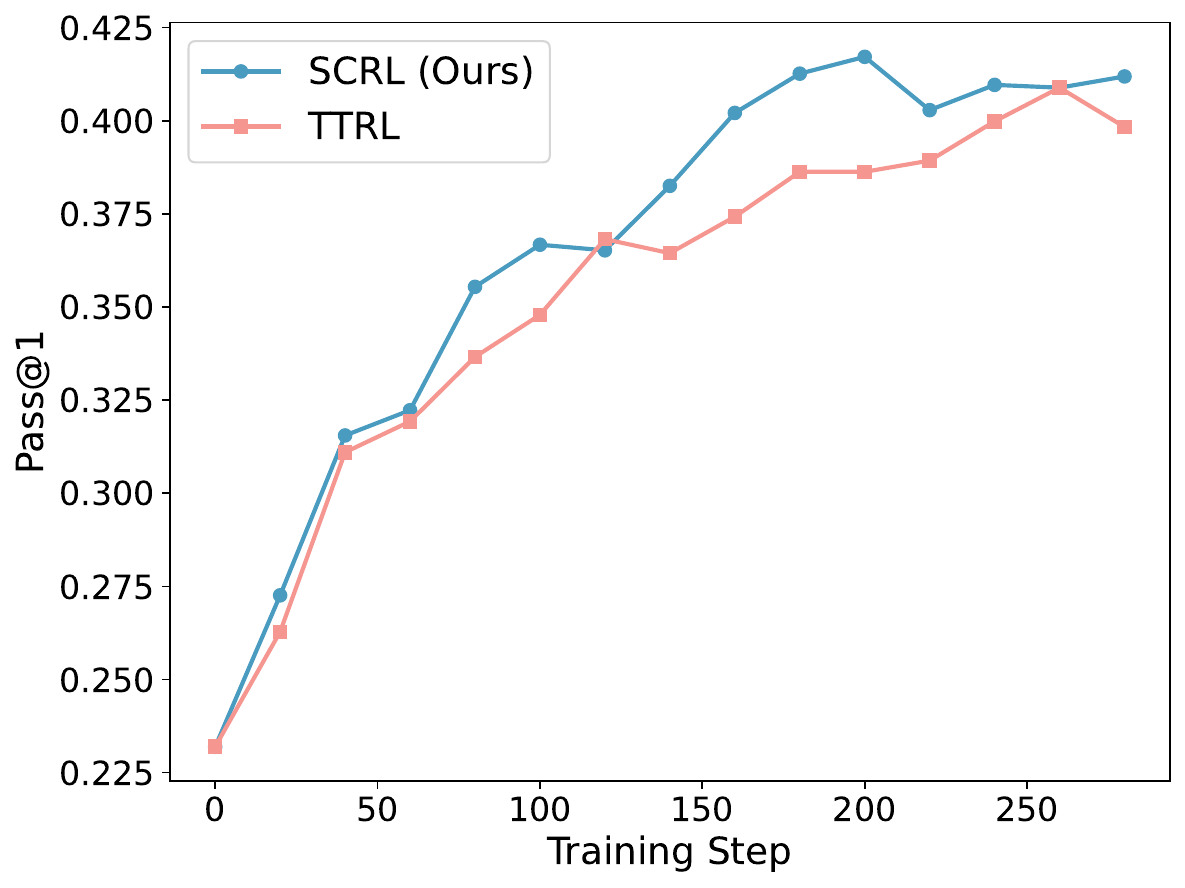}
    \caption{AMC}
    \label{fig:b}
  \end{subfigure}\hfill
  \begin{subfigure}[t]{0.32\textwidth}
    \centering
    \includegraphics[width=\linewidth]{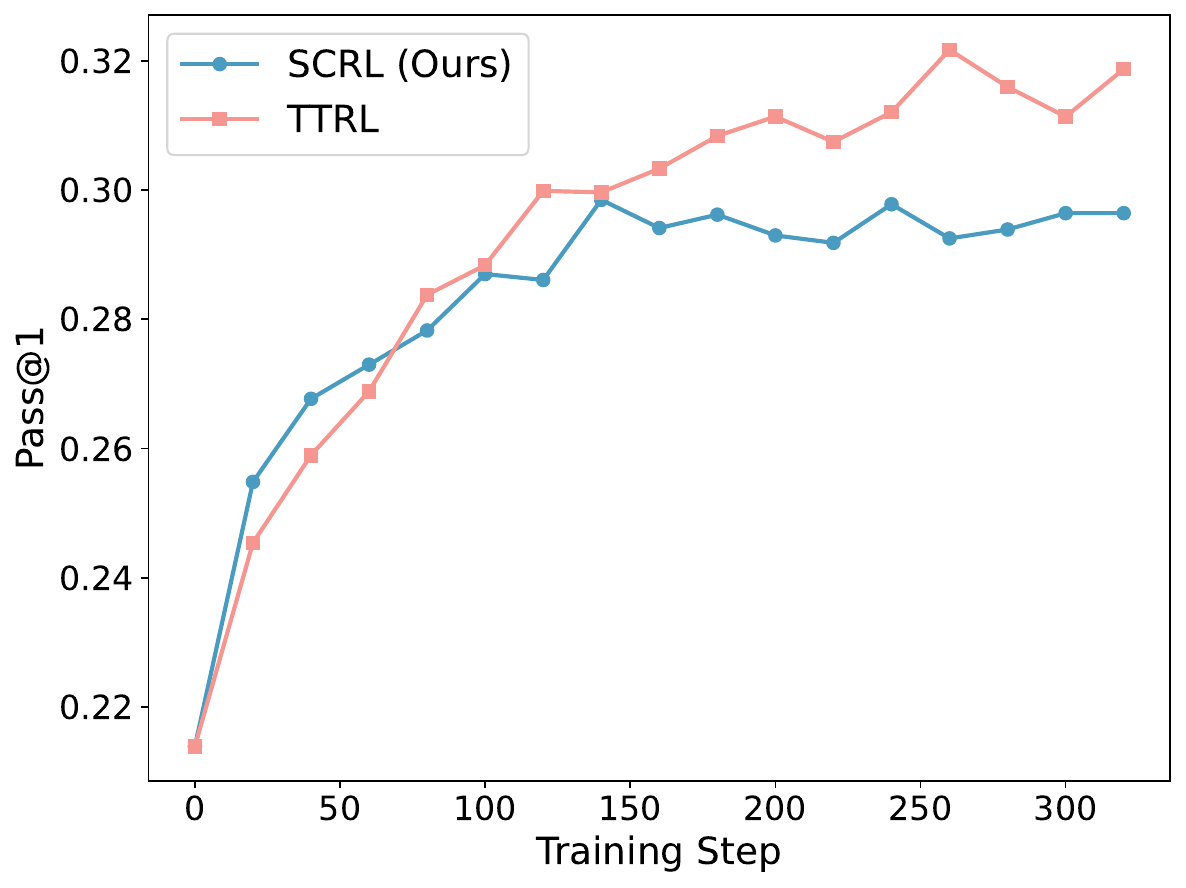}
    \caption{Minerva}
    \label{fig:c}
  \end{subfigure}
  \caption{Training dynamics of SCRL and TTRL on Qwen2.5-3B across three mathematical benchmarks.}
  \label{fig:qwen25-3b-train-dynamic}
\end{figure*}
\begin{figure*}[!htbp]
  \centering
  \begin{subfigure}[t]{0.32\textwidth}
    \centering
    \includegraphics[width=\linewidth]{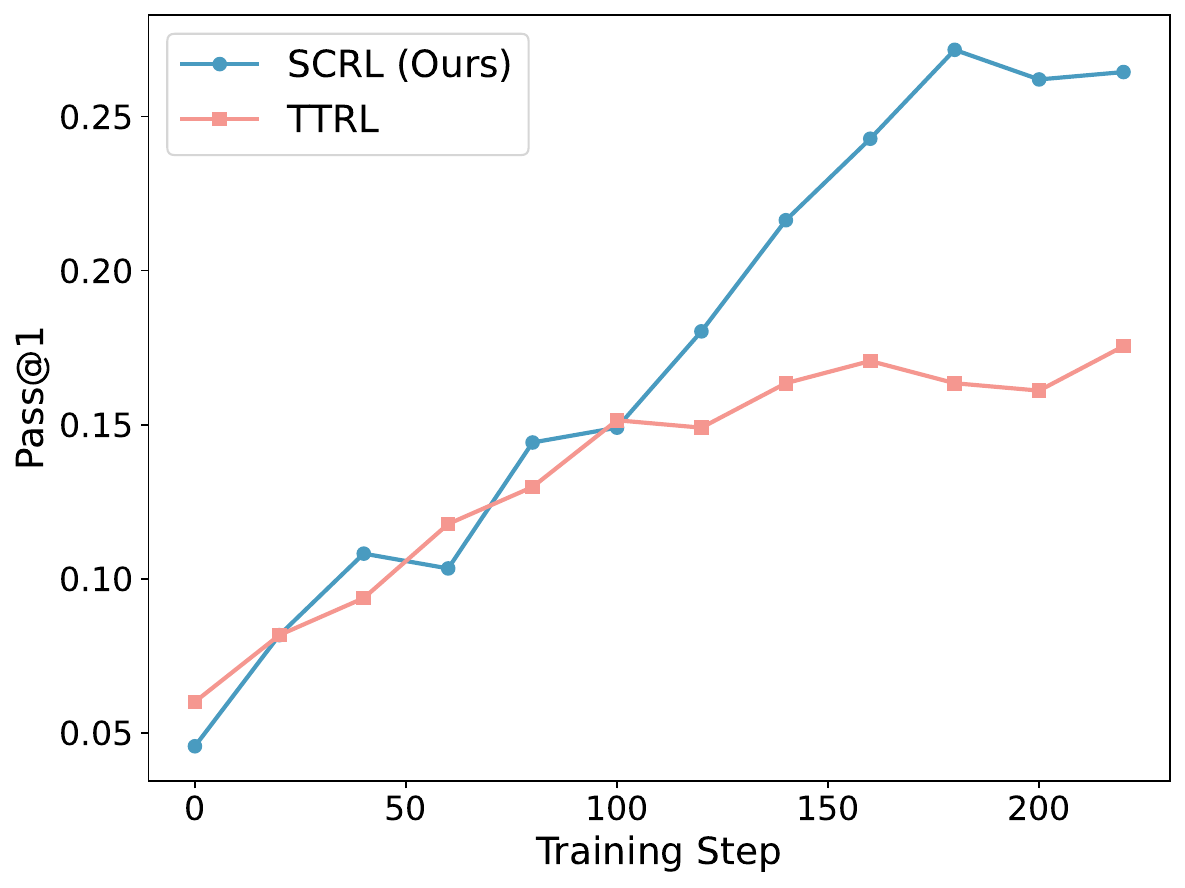}
    \caption{AIME25}
    \label{fig:a2}
  \end{subfigure}\hfill
  \begin{subfigure}[t]{0.32\textwidth}
    \centering
    \includegraphics[width=\linewidth]{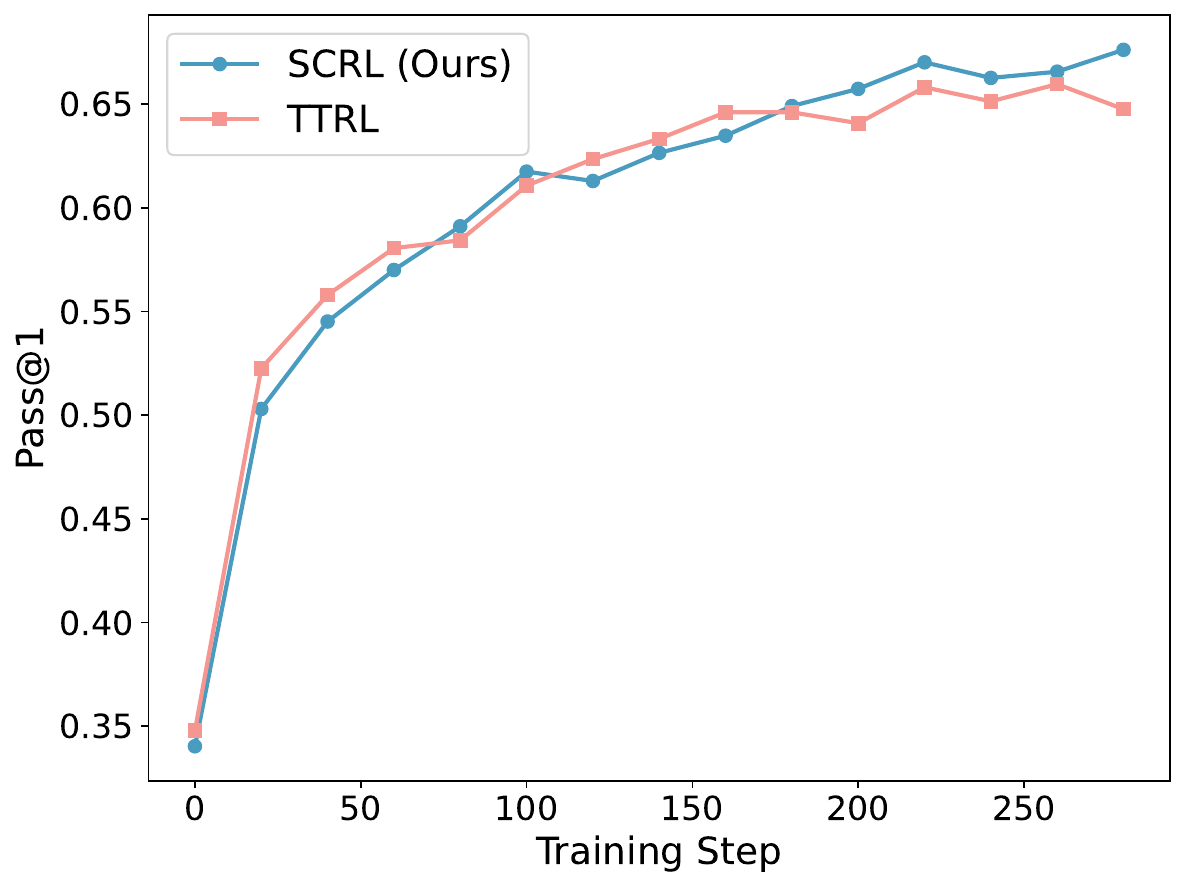}
    \caption{AMC}
    \label{fig:b2}
  \end{subfigure}\hfill
  \begin{subfigure}[t]{0.32\textwidth}
    \centering
    \includegraphics[width=\linewidth]{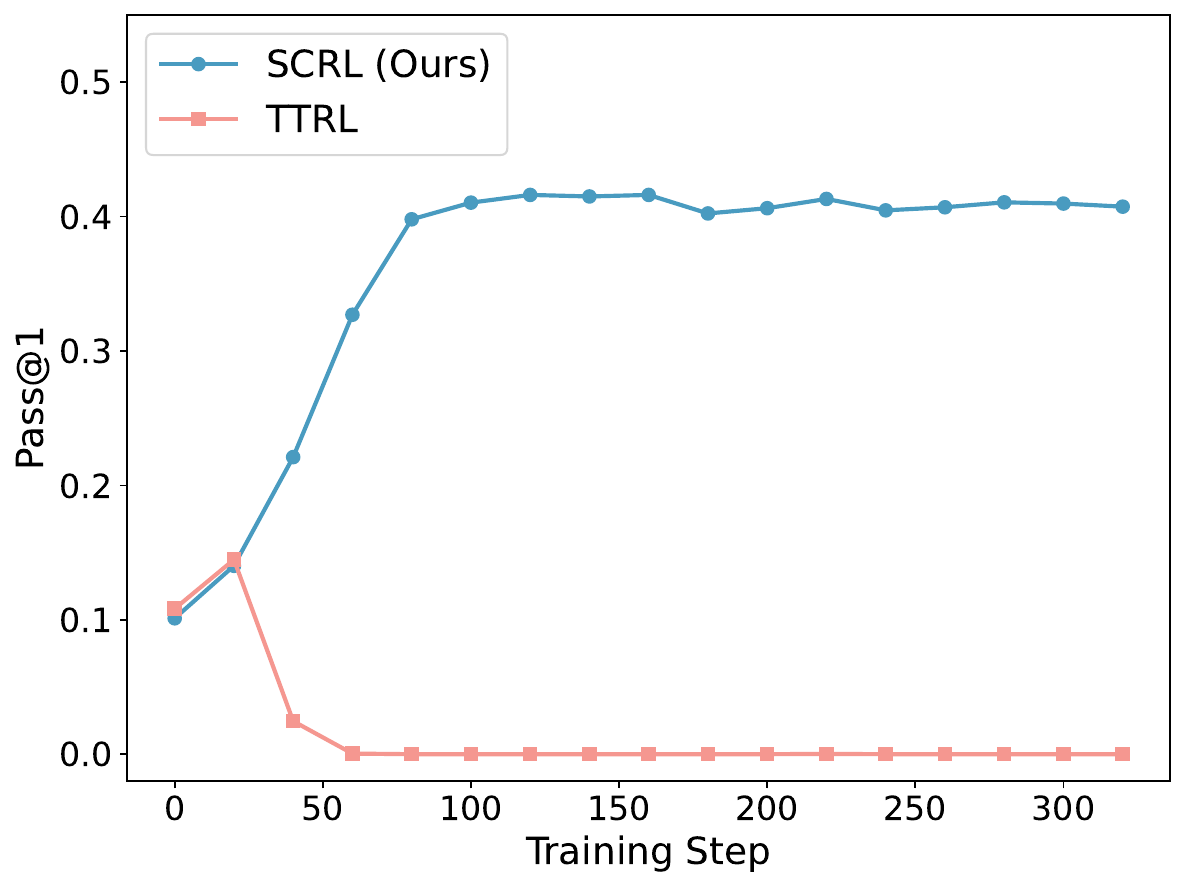}
    \caption{Minerva}
    \label{fig:c2}
  \end{subfigure}

  \caption{Training dynamics of SCRL and TTRL on Qwen2.5-Math-7B across three mathematical benchmarks.}
  \label{fig:qwen25math-train-dynamic}
\end{figure*}

\subsection{Benchmarks}
Our evaluation suite comprises six challenging reasoning datasets that span different difficulty levels and problem-solving domains:
\begin{itemize}
\item \textbf{AIME24 and AIME25} \citep{li2024numinamath}: The official problem sets from the 2024 and 2025 American Invitational Mathematics Examination, respectively.
\item \textbf{AMC} \citep{li2024numinamath}: A collection of problems from the American Mathematics Competitions, covering core high-school mathematical domains.
\item \textbf{MATH-500} \citep{hendrycks2021measuring}: A 500-problem subset sampled from the MATH dataset, covering mathematical reasoning problems across algebra, geometry, number theory, and combinatorics.
\item \textbf{Minerva} \citep{team2024qwen2}: The mathematics subset from the Minerva quantitative reasoning benchmark.
\item \textbf{GPQA} \citep{rein2024gpqa}: A challenging dataset of 448 graduate-level multiple-choice questions written by domain experts in biology, physics, and chemistry.
\end{itemize}

\subsection{Training Episode}
For the MATH-500 dataset using Qwen2.5-3B with 32 candidate responses and 16 training samples, the TTRL method experiences training collapse due to excessive optimization steps when the training episode is set to 10. Consequently, we set the training episode to 4 for this specific configuration to ensure fair comparison. All other settings remain consistent with the descriptions in the Section~\ref{experiment_setup}.
\subsection{Training Hyperparameter}
Our training hyperparameters are shown in Table~\ref{tab:parameter}.

\section{Additional Results}
Figures~\ref{fig:qwen25-3b-train-dynamic} and~\ref{fig:qwen25math-train-dynamic} show the pass@1 accuracy (\%) throughout training.

\section{Analysis of Consensus Margin and Entropy Penalty}
\label{sec:margin_analysis}
We conduct an analysis of $\tau_{\text{marg}}$ and $\lambda_H$ using Qwen2.5-Math-7B with 64 candidate responses and 32 training samples. Table~\ref{tab:hyperment2} presents the results across different threshold configurations. 

The entropy penalty weight $\lambda_H$ controls the strength of uncertainty penalization in our dynamic reward shaping mechanism. Setting $\lambda_H=0.1$ achieves the optimal performance across both benchmarks. Compared to $\lambda_H=0$, introducing a moderate penalty provides a significant performance gain, confirming that discouraging high-uncertainty trajectories helps the policy avoid converging to unstable solutions. However, increasing the coefficient leads to a performance degradation, which suggests that moderate entropy regularization effectively balances the trade-off between encouraging confident reasoning and maintaining sufficient exploration capacity.

The margin threshold $\tau_{\text{marg}}$ enforces separation between the top-ranked answer and alternatives, preventing unreliable majorities from being reinforced. The results highlight a distinct trade-off based on problem difficulty. On AIME25, increasing $\tau_{\text{marg}}$ from 0 to 0.125 achieves substantial improvements, with further increase to 0.25 maintaining strong performance at 22.6\%. This demonstrates that for challenging problems where answer distributions are highly dispersed, strict margin requirements are crucial to filter out false-positive consensus. Conversely, on the relatively easier AMC dataset, a lower margin of $\tau_{\text{marg}}=0.0625$ achieves higher performance, which suggests that on problems where the model exhibits sufficient accuracy, overly strict margins may unnecessarily discard valid positive signals. Our configuration in main experiments demonstrates the most robust generalization across varying difficulty levels, effectively mitigating label noise on hard tasks while maintaining supervision frequency on easier ones.
\begin{table}[h]
\centering
\begin{tabular}{ll|cc}
\hline
$\tau_{\text{marg}}$ &$\lambda_H$ &\textbf{AIME25}& \textbf{AMC}\\
\hline
0.125&0&19.0&64.4\\
0.125&0.2&19.0&63.7\\
0.125&0.5&18.5&66.5\\
0&0.1&18.8&66.5\\
0.0625&0.1&18.8& 68.8\\
0.25&0.1&22.6&65.6\\
\rowcolor[HTML]{D7F6FF}
\textbf{0.125}&\textbf{0.1}&\textbf{22.8}&\textbf{68.5}\\
\hline
\end{tabular}
\caption{Parameter analysis of $\tau_{\text{marg}}$ and $\lambda_H$ on Qwen2.5-Math-7B. The table reports pass@1 accuracy (\%). Bold shows the configuration in main experiments.}
\label{tab:hyperment2}
\vspace{-0.5\baselineskip}
\end{table}

\end{document}